\RequirePackage{fix-cm}
\documentclass[twocolumn]{svjour3}          
\smartqed  

\usepackage{tabularx}
\usepackage{cite}
\usepackage[pdftex]{graphicx}
\usepackage{ragged2e}
\usepackage[tight,footnotesize]{subfigure}
\usepackage{rotating}
\usepackage{graphicx}
\usepackage{amsmath,amssymb} 
\usepackage{array}
\usepackage{multirow}
\usepackage{colortbl}
\usepackage{amsfonts}
\usepackage{pifont}
\usepackage{xspace}
\usepackage{etoolbox}
\usepackage{overpic}
\usepackage{color}
\usepackage{microtype}
\usepackage{bm}
 \usepackage{booktabs}

\newcommand{\orcid}[1]{\href{https://orcid.org/#1}{\includegraphics[width=10pt]{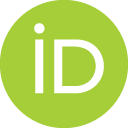}}}

\def\etal{{\em et al}}

\usepackage{hyperref}
\hypersetup{breaklinks=true,citecolor=blue, colorlinks}

\graphicspath{{./Imgs/}}
\DeclareGraphicsExtensions{.pdf,.jpg,.png}

\usepackage{silence}
\journalname{Research Article}

\begin{document}

\title{FusionMamba: Dynamic Feature Enhancement for Multimodal Image Fusion with Mamba}

\titlerunning{FusionMamba}        

\author{Xinyu Xie \orcid{(0000-0001-5039-3349}\and
  Yawen Cui \orcid{0000-0002-9337-687X} \and 
  Tao Tan \orcid{0000-0001-5403-0887} \and
  Xubin Zheng \orcid{0000s-0003-2322-857X} \and
  Zitong Yu \orcid{0000-0003-0422-6616} \\
}

\authorrunning{Xinyu Xie \etal} 

\institute{
Xinyu Xie is the first author, and Zitong Yu is the corresponding author of this paper.
Xinyu Xie, Xubin Zheng and Zitong Yu are with the Department of Computer Science, Great Bay University, Dongguan, Guangdong, 523000, China (Email: xbzheng@gbu.edu.cn, yuzitong@gbu.edu.cn). 
Yawen Cui is with the Department of Computer Science, The Hong Kong Polytechnic University, Hong Kong, China (Email: yawen.cui@polyu.edu.hk). 
Xinyu Xie and Tao Tan are with the Macao Polytechnic University, Macao, China (Email: taotanjs@gmail.com).
}

\date{Received: date / Accepted: date}

\maketitle

\begin{abstract}
Multimodal image fusion aims to integrate information from different imaging techniques to produce a comprehensive, detail-rich single image for downstream vision tasks. Existing methods based on local convolutional neural networks (CNNs) struggle to capture global features efficiently, while Transformer-based models are computationally expensive, although they excel at global modeling. Mamba addresses these limitations by leveraging selective structured state space models (S4) to effectively handle long-range dependencies while maintaining linear complexity. In this paper, we propose FusionMamba, a novel dynamic feature enhancement framework that aims to overcome the challenges faced by CNNs and Vision Transformers (ViTs) in computer vision tasks. The framework improves the visual state-space model Mamba by integrating dynamic convolution and channel attention mechanisms, which not only retains its powerful global feature modeling capability, but also greatly reduces redundancy and enhances the expressiveness of local features. In addition, we have developed a new module called the dynamic feature fusion module (DFFM). It combines the dynamic feature enhancement module (DFEM) for texture enhancement and disparity perception with the cross-modal fusion Mamba module (CMFM), which focuses on enhancing the inter-modal correlation while suppressing redundant information. Experiments show that FusionMamba achieves state-of-the-art performance in a variety of multimodal image fusion tasks as well as downstream experiments, demonstrating its broad applicability and superiority.

\keywords{Multimodal \and Image fusion \and Feature enhancement \and Mamba}

\end{abstract}

\section{Introduction}
\begin{figure*}[!t]
    \centering
    \includegraphics[width=160mm]{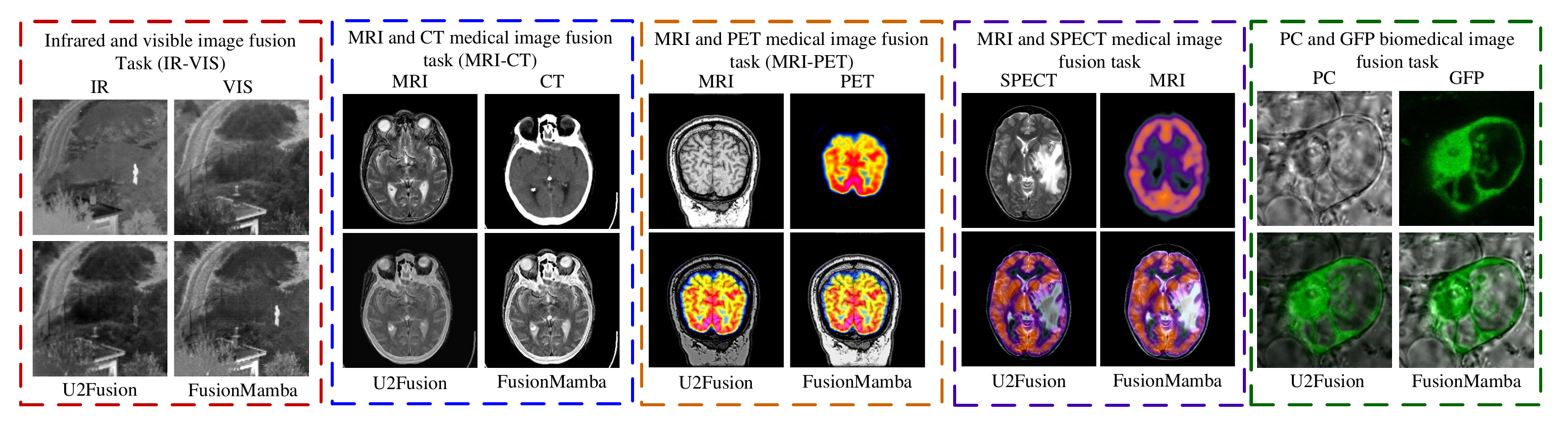} 
    \caption{Illustration of qualitative and quantitative results of multimodal image fusion. Qualitative visualization between classical U2Fusion \cite{U2Fusion} and our FusionMamba is shown in the second row while the sub-figures on the first row are source image pairs.}
    \label{fig:FusionMamba_1}
    \vspace{-0.4em} 
\end{figure*}

Multimodal image fusion is an advanced technique that integrates information from different imaging modalities into a single image \cite{U2Fusion}. By combining the unique advantages of each modality, this method provides more comprehensive and accurate image data \cite{In1}. It demonstrates extensive application potential and significant benefits in fields such as medical imaging, remote sensing, and computer vision \cite{Me1, yu2021searching, ye2024cat, yu2023rethinking}. Multimodal image fusion effectively overcomes the inherent limitations of single imaging techniques, significantly improving diagnostic accuracy, providing analytical insights, and enhancing both visualization and feature extraction precision. In security monitoring and autonomous driving, the fusion of infrared and visible images (IVF) combines the advantages of infrared imaging in low-light and obstructed conditions with the detailed presentation capabilities of visible light images under standard lighting. This fusion greatly enhances target detection and recognition abilities, and improves system stability and reliability in complex environments \cite{SwinFusion}. Thus, multimodal image fusion technology not only advances precision medicine, but also plays a crucial role in enhancing security technologies and optimizing autonomous driving perception systems \cite{In2}. In medical imaging, multimodal medical image fusion (MIF) integrates modalities such as positron emission tomography (PET), computed tomography (CT), and magnetic resonance imaging (MRI) to provide richer information on tissues and lesions. This integration aids in improving diagnostic accuracy, reducing the risk of misdiagnosis and missed diagnoses, and optimizing treatment decisions. It enhances the precision of lesion localization and the detail of lesion assessment, providing a more scientific basis for clinical treatment planning \cite{Impoptantnet}. A schematic illustration of these types of multimodal image fusion tasks is exhibited in Figure. \ref{fig:FusionMamba_1}.

In recent years, there has been a significant increase in the use of deep learning for multimodal image fusion, primarily employing convolutional neural networks (CNNs) \cite{U2Fusion,RoadScene,Densefuse,IFCNN} and Transformer architectures \cite{IFT,SwinFusion,MATR,MRSCFusion,DATFuse} for feature extraction and reconstruction. However, these models face limitations. Specifically, the utilization of static convolutional layers in these fusion methods results in uniform traversal of the entire image, limiting their ability to capture global contextual information due to constrained receptive fields \cite{SwinFusion}. Moreover, the indiscriminate processing of each pixel by static convolutional layers overlooks nuanced spatial differences across various positions. Because image fusion tasks require a model to process diverse input data immediately, current convolution-based techniques struggle to extract features effectively from different modalities, leading to diminished fusion performance.  Transformer-based models excel in global modeling but suffer from quadratic complexity concerning image size due to their self-attention mechanism \cite{attention}, imposing substantial computational overhead.   Additionally, Transformers exhibit lower precision in capturing local features compared to CNNs. Although certain fusion models \cite{MRSCFusion,SwinFusion,IFT,CDDFuse} adopt a hybrid approach by amalgamating convolutional and Transformer layers to leverage their respective strengths and mitigate weaknesses, computational overhead remains a significant concern.

Image fusion is an essential image enhancement technique \cite{SwinFusion}. Researchers often employ three main feature fusion strategies. First, some methods use element-wise operations like addition, multiplication, or concatenation \cite{IFCNN,Nestfuse} to merge features from different modalities.  However, these approaches overlook inter-modal relationships, leading to compromised fusion performance.  Second, there are approaches that focus on deeper feature extraction \cite{IFT,MRSCFusion}, but they still lack effective inter-modal interaction and texture detail.  Last, specific techniques utilize cross-modal fusion \cite{SwinFusion}, including convolution-based and Transformer-based methods, which partially address feature interaction but have their limitations. Overall, current fusion methods struggle to make optimal use of modal features and highlight crucial information, indicating a need for improved modal connections and emphasis on key image details.

The advancement of Mamba \cite{Mamba} represents a promising avenue for achieving a balance between global receptive fields and computational efficiency. By formalizing the discretized state-space equations of Mamba into recursive form and incorporating specially designed structured reparameterization \cite{MambaIR}, it becomes capable of simulating very long-range dependencies. Moreover, the implementation of a parallel scanning algorithm \cite{Vm-unet} enables Mamba to process each token in a parallel manner, thereby facilitating efficient training on modern hardware like GPUs \cite{EVMamba}. These encouraging characteristics inspire us to further explore Mamba's potential for efficient long-range modeling within image fusion networks.

In response to the above issues, our study designs a new dynamic feature fusion model with Mamba for multimodal image fusion, which aims to better explore intra-modal and inter-modal features while dynamically enhancing the detailed texture information of the source images and the specificity information of each modality. Specifically, for the feature extraction and reconstruction part of the network, we design a Mamba model suitable for fusion tasks by integrating the visual state space model with dynamic convolution and channel attention, which not only maintains Mamba's performance and global modeling capabilities, but also reduces channel redundancy and enhances local feature extraction capabilities. For the feature fusion part, we design a dynamic feature fusion module  (DFFM) comprising a dynamic feature enhancement module (DFEM) and a cross modality fusion Mamba module (CMFM). Unlike previous feature fusion methods, this fusion module dynamically enhances the detailed texture information and differential information in source images and facilitates better information interaction between modalities. The DFEM, composed of dynamic differential convolution and dynamic difference-sensing attention, is used for adaptive feature enhancement. The DFEM dynamically enhances critical information by establishing connections between input features from different modalities. We design a CMFM to effectively mine correlation information between modalities. Our network architecture adopts a Unet \cite{Unet} multi-layer structure, realizing an efficient and universal image fusion framework. Experimental results demonstrate that our proposed method outperforms state-of-the-art image fusion methods across various evaluation metrics on multiple multimodal baseline datasets, including infrared and visible (IR-VIS) image fusion, CT-MRI image fusion, PET-MRI image fusion, SPECT-MRI image fusion, and green fluorescent protein and phase contrast (GFP-PC) image fusion.

In summary, our contributions are summarized as follows.

1) We design a novel dynamic feature-enhanced Mamba image fusion model, offering an effective alternative to methods based on CNNs and Transformers.

2) We propose the dynamic visual state space (DVSS) block, which enhances the efficiency of the standard Mamba model by dynamically enhancing local features and reducing channel redundancy. This enhancement strengthens its modeling and feature extraction capabilities.

3) The feature fusion module extracts key information from source images and explores relationships between different modalities.  It includes a dynamic feature enhancement module to enhance fine-grained texture features and detect differential features, along with a Mamba cross modality fusion module to effectively explore inter-modal correlations.

4) We develop an efficient and versatile image fusion framework, achieving leading performance in various image fusion tasks, including  IR-VIS image fusion, multimodal medical image fusion, and biomedical image fusion.

\section{Related Work}
\subsection{Deep Multimodal Image Fusion}
In recent years, with the increasing application of deep learning image vision, researchers have introduced deep learning methods into the area of image fusion. These frameworks mainly include CNNs, generative adversarial networks (GANs), autoencoder networks (AEs), Transformer-based networks, and other methods.

Early deep learning-based approaches typically adopted CNN frameworks and used simple fusion rules such as element-wise addition, averaging, or multiplication. For example, Li et al. \cite{Densefuse} proposed DenseFuse, a pre-trained fusion model with CNN as the backbone network. The model uses dense connections in the encoder layer to achieve effective fusion results. Liu et al. \cite{liu2017multi} introduced a CNN model that utilizes a Laplacian pyramid framework for activity-level measurement and fusion rule generation. Liang et al. \cite{MCFnet} proposed an end-to-end multi-layer concatenation fusion network. Subsequently, convolution-based methods further improved fusion performance by designing loss functions and fusion strategies. PMGI \cite{PMGI} preserves the relationship between gradients and intensities across various image fusion tasks through a unified loss function. Zhang et al. \cite{U2Fusion} enhanced fusion performance using a squeeze-and-decomposition network and adaptive decision blocks. CoConet \cite{CoConet} introduced a convolution-based coupled contrastive learning approach for general image fusion. NestFuse \cite{Nestfuse} proposed by Li et al. and RFN-Nest \cite{RFN} proposed by Li et al. introduced multi-scale architectures, nested connections, and convolutional fusion layers to extract comprehensive features. SeAFusion \cite{SeAFusion} designed by Tang et al. utilized a dense gradient convolutional layer and a segmentation network to enhance feature extraction for finer-grained details. PIAFusion \cite{PIAFusion} proposed an illumination-aware progressive fusion network based on CNN. Although these methods can avoid the issues of traditional algorithms being limited to single tasks and computational complexity, they are still constrained by the inherent local receptive fields of convolutions, which affects fusion performance. Ma et al.  \cite{FusionGAN} introduced FusionGAN, the first GAN-based method for image fusion. Li et al. \cite{AttentionFGAN} improved FusionGAN by embedding a multi-scale attention mechanism in both the generator and discriminator, leading to the development of AttentionFGAN. Li et al. \cite{MgANFuse} also presented MgANFuse, a GAN-based model featuring a multi-grained attention module within an AE framework for fusing IR and VIS images. Despite these methods enhancing feature extraction through advanced network architectures, they still rely on manually crafted fusion strategies, potentially limiting their performance.

With the rise of Transformers, the issue of CNNs' inability to extract global features has been addressed. Tang et al. \cite{DATFuse} used DATFuse, the spatial and channel attention mechanism for global and local feature extraction. Chang et al. \cite{AFT} introduced AFT, an adaptive fusion Transformer network that focuses on adaptive information localization in IR and VIS image fusion tasks using spatial domain-based modules. Since single Transformer-based fusion networks may not be as accurate in local feature extraction, some fusion methods that combine the advantages of CNNs and Transformers have been proposed. Tang et al. \cite{YDTR} proposed YDTR, a Y-shaped network combining CNN and Transformer architectures to capture both local and global information, aiming to preserve details more effectively. Xie et al. \cite{MRSCFusion} combined CNN and Swin Transformer \cite{Swin} for effective feature learning. MATR designed by Tang et al. \cite{MATR} developed an adaptive multi-scale Transformer for multimodal medical image fusion, while MACTFusion designed by Xie et al. \cite{MACTFusion} introduced a cross-modal fusion Transformer framework. VS et al. \cite{IFT} utilized IFT, a Transformer-based multi-scale fusion strategy to effectively integrate local and long-range dependency information. Feng et al. \cite{SwinFusion} developed a multi-task fusion model that employs cross-domain long-range learning to effectively integrate dependencies both within and between domains. By combining CNNs and Transformers, they effectively capture local information and integrate global complementary features. Zhao et al. \cite{CDDFuse} proposed a feature decomposition fusion model that uses CNN and Transformer to extract modality-specific and shared features for cross-modal feature modeling and decomposition. Although these Transformer-based frameworks can significantly improve fusion performance, their self-attention mechanisms lead to high computational costs.

\subsection{Mamba}
State space models (SSMs) \cite{SSM} have become a competitive backbone in deep learning, originating from classic control theory and offering linear scalability with sequence length for long-range dependency modeling. Structured state space sequence models (S4) and Mamba \cite{VMamba}, both rely on a classical continuous system that maps a 1D input function or sequence, denoted as \(x(t) \in \mathbb{R}^N\), through intermediate implicit states \(h(t) \in \mathbb{R}^N\) to an output \(y(t) \in \mathbb{R}^N\). N represents the state size. SSMs can be represented as the following linear ordinary differential equation: 
\begin{equation}
\begin{gathered}
h^{\prime}(t) = \bm{A}h(t) + \bm{B}x(t),\\
y(t)  = \bm{C}h(t) + \bm{D}x(t),
\end{gathered}
\end{equation}
where \(\bm{A} \in \mathbb{R}^{N \times N}\) represents the state matrix, while \(\bm{B} \in \mathbb{R}^{N \times 1}\), \(\bm{C} \in \mathbb{R}^{N \times 1}\), and \(\bm{D} \in \mathbb{R}^{N}\) denote the projection parameters.
After that, the discretization process is typically adopted into practical deep learning algorithms. Specifically, denote \(\Delta\) as the timescale parameter to transform the continuous parameters \(\bm{A}\), \(\bm{B}\) to discrete parameters \(\overline{\bm{A}}\), \(\overline{\bm{B}}\). The commonly used method for discretization is the zero-order hold rule, which is defined as follows:
\begin{equation}
\begin{gathered}
\overline{\bm{A}} = \exp\left( \Delta\bm{A} \right),\\
\overline{\bm{B}} = \left( \Delta \bm{A} \right)^{-1} \left( \exp\left( \Delta \bm{A} \right) - \bm{I} \right) \cdot \Delta \bm{B}.
\end{gathered}
\end{equation}
After the discretization, the discretized version of Eq. (1) with step size \(\Delta\)
can be rewritten in the following RNN form:
\begin{equation}
\begin{gathered}
h_{k}=\overline{\bm{A}}h_{k-1}+\overline{\bm{B}}x_{k},\\
y_{k}=\bm{C}h_{k}+\bm{D}x_{k}.
\end{gathered}
\end{equation}
\vspace{-1.5em}

Furthermore, the Eq. (3) can also be mathematically equivalently transformed into the following CNN form:
\begin{equation}
\begin{gathered}
\overline{\bm{K}}\overset{\Delta }{=}\left ( \bm{C}\overline{\bm{B}},\bm{C}\overline{\bm{AB}},\cdots,\bm{C}\overline{\bm{A}}^{L-1}\overline{\bm{B}} \right ),\\
\bm{y}=\bm{x}\circledast \overline{\bm{K}},
\end{gathered}
\end{equation}
where \(\circledast\) denotes convolution operation, \(\overline{\bm{K}}\in \mathbb{R}^{L}\) is a structured convolution kernel and \(\textit{L}\) denotes the length of the input sequence \(\bm{x}\).

Mamba has significantly advanced natural language tasks, surpassing traditional Transformers, with its data-dependent mechanisms, efficient hardware, and superior language processing. Expanding beyond language tasks, Mamba has also been applied successfully to vision tasks like image classification, video understanding, and biomedical image segmentation. This success has spurred a wave of research focusing on adapting Mamba-based models for specialized vision applications, including medical image segmentation with adaptations like VM-Unet \cite{Vm-unet}. Additionally, Mamba has been integrated into graph representation tasks, enhancing graph embeddings and processing capabilities through models. Mamba's versatility and efficiency make it a compelling choice for a wide range of applications, from language processing to computer vision representation tasks \cite{MambaIR,EVMamba}. FMamba \cite{Dehazing} incorporates a multi-scale progressive fusion module designed for image dehazing. In the field of image fusion, Dong et al. \cite{MD}  proposed a cross-modal fusion framework based on Mamba for object detection. Mambadfuse \cite{Mambadfuse} designed a structure that combines ResNet with Mamba for image fusion.

\section{Methodology}
\subsection{Overview}
\begin{figure*}[t]
  \centering
  \includegraphics[width=\textwidth]{ 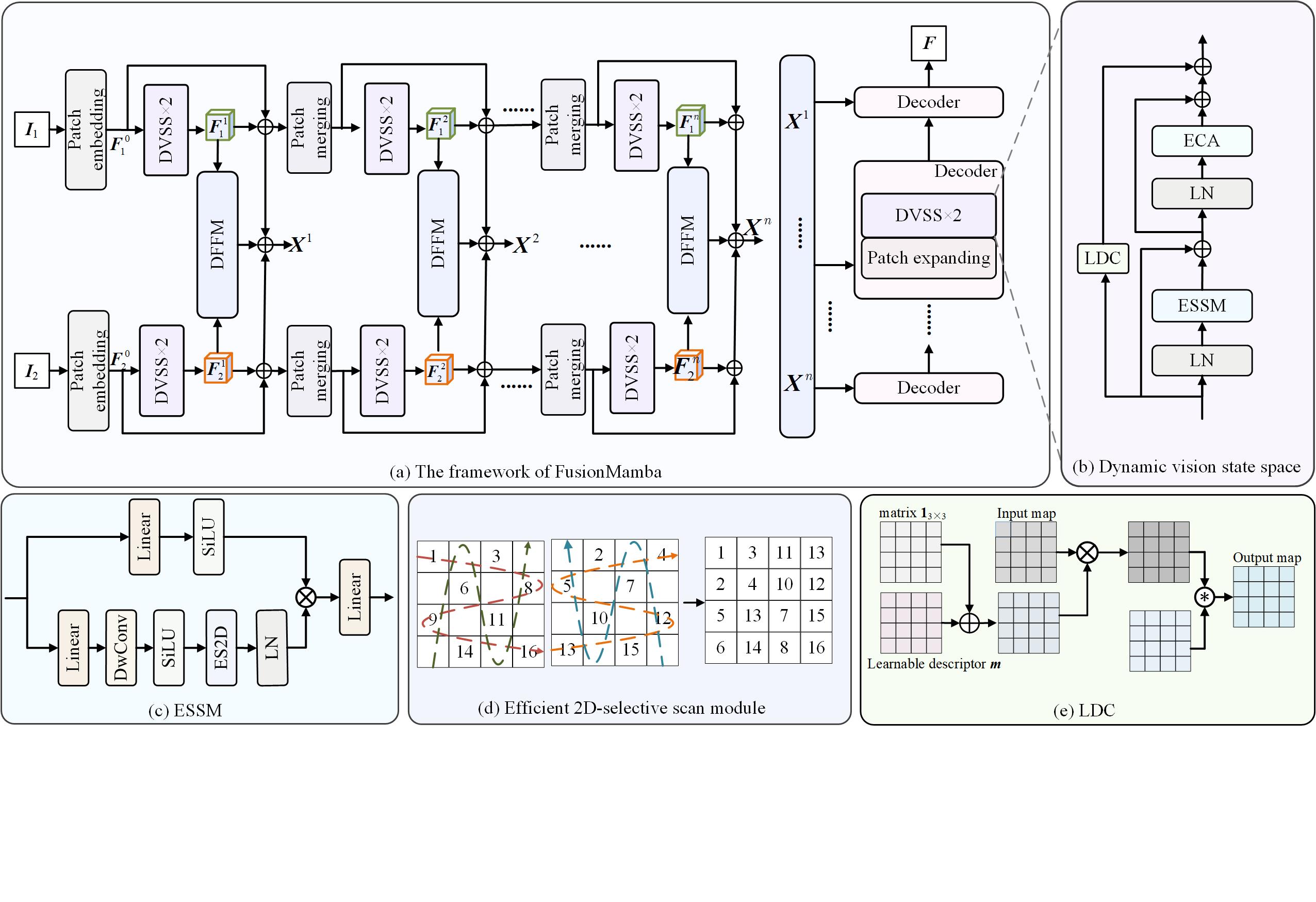}
  \vspace{-8em}
  \caption{Overview of the framework. FusionMamba network receives two images of different modes as inputs. These images undergo multi-layer feature extraction and dynamic feature enhancement fusion through the fusion module, resulting in fusion features that include difference and texture enhancement. Finally, the module reconstructs the fusion result. I1, I2: different source images; F: fused image; DFFM: dynamic feature fusion module; DVSS: dynamic vision state space. LDC: learnable descriptive convolution; ECA: efficient channel attention; LN: LayerNorm; ESSM: efficient state space module. Linear: linear function; DwConv: depthwise separable convolution; SiLU: SiLU activation function; ES2D: the efficient 2D scanning.}
  \label{fig:FusionMamba}
  \vspace{-0.3em}
\end{figure*}

Our FusionMamba framework encompasses three critical components of the fusion process: feature extraction, feature fusion, and feature reconstruction. Inspired by VM-Unet \cite{Vm-unet}, our network architecture (in Figure. \ref{fig:FusionMamba}) is based on the Unet framework, which effectively extracts deeper features. In this section, we introduce the overall framework and provide detailed explanations of the key modules in the following subsections.

The network consists of four layers (refer to VM-unet \cite{Vm-unet}). In the first layer, we partition the input image into non-overlapping patches of size $4 \times 4$, and then map these patches to a feature space of dimension $C$, where $C$ is 96 by default. The size of the embedded image after mapping is $\frac{H}{4} \times \frac{W}{4} \times C$. In the first three layers, we apply a patch merging layer to reduce the height and width of the input features while increasing the number of channels. A patch expanding operation is utilized to decrease the number of feature channels and increase the height and width. The number of channels for each layer is [C, 2C, 4C, 8C]. As shown in Figure. \ref{fig:FusionMamba}(a), the feature extraction and reconstruction stages in each layer utilize the designed Dynamic Visual State Space (DVSS) module. The number of DVSS modules in each layer is [2, 2, 9, 2]. During the feature fusion stage, we employ the dynamic feature fusion module (DFFM). Each layer of the DFFM contains two DFEMs and one CMFM. Consequently, in the following sections, we will focus on the designed DVSS modules, DFEM, and CMFM.
\subsection{Dynamic Vision State Space Module}
Inspired by EVMamba \cite{EVMamba}, we propose the DVSS module as a modification of the SSM block for image fusion process.  In Figure. \ref{fig:FusionMamba}(b), starting with the input deep feature, we initially apply LayerNorm (LN) followed by the efficient state space module (ESSM) 
 \cite{EVMamba}, as shown in Figure.\ref{fig:FusionMamba}(c), to capture spatial long-term dependencies. Since SSMs process flattened feature maps as 1D token sequences, the chosen flattening strategy significantly affects the number of neighboring pixels in the sequence.  For example, when using the four-direction unfolding strategy, only four nearest neighbors are accessible to the anchor pixel. Specially,  efficient 2D scan (ES2D, in Figure. \ref{fig:FusionMamba}(d)) \cite{EVMamba} downscales 2D-Selective-Scan (SS2D) using skipping sampling and combines processed patches for global feature extraction.  Consequently, some spatially close pixels in the 2D feature map become distant from each other in the 1D token sequence, potentially resulting in local pixel forgetting.  To address this issue, we introduce an additional dynamic local convolution (LDC) \cite{LDC} after ES2D \cite{EVMamba} to restore neighborhood similarity. The LDC, in Figure. \ref{fig:FusionMamba}(d)) effectively learns intricate textural features, making it highly suitable in here. we employ LayerNorm for normalization and then utilize LDC \cite{LDC} to compensate for local features.
The LDC is calculated using Eq. (5):
\begin{equation}
\begin{split}
\mathrm{LDC}\left (p\right ) &= \bm{w}(p) \cdot \left(\bm{f}(p) \odot \left((1-\varepsilon) \cdot \mathbf{1}_{3 \times 3}(p) + \varepsilon \cdot \bm{m}(p)\right)\right) \\
&= (1-\varepsilon) \cdot (\bm{w}(p) \cdot \bm{f}(p)) \\
& \quad + \varepsilon \cdot (\bm{w}(p) \cdot (\bm{f}(p) \odot \bm{m}(p))) \\
&= (1-\varepsilon) \underbrace{\sum_{p_{n} \in \mathcal{R}} \bm{w}\left(p_{n}\right) \cdot \bm{f}\left(p+p_{n}\right)}_{\text{vanilla convolution}} \\
& \quad + \varepsilon \underbrace{\sum_{p_{n} \in \mathcal{R}} \bm{w}\left(p_{n}\right) \cdot \left(\bm{f}\left(p+p_{n}\right) \cdot \bm{m}\left(p_{n}\right)\right)}_{\text{learnable descriptive convolution}},
\end{split}
\end{equation}
where \( \cdot \) denotes the convolution operation, \( p \) represents the current pixel location, and \( n \) refers to the location of neighboring pixels in the local neighborhood \( \mathcal{R} \). \(\bm{w}\) denotes the convolution kernel, and \(\bm{f}\) represents the input map. The symbol \( \odot \) indicates element-wise multiplication. \( \varepsilon \) is a learnable parameter that controls the balance between the contributions of two convolutional operations. As shown in Figure. \ref{fig:FusionMamba}(e), the base matrix \( \mathbf{1}_{3 \times 3} \) is an all-ones matrix, which ensures that the standard convolution operation is preserved in the LDC. Meanwhile, the matrix \(\bm{m} \) is also initialized as \( \mathbf{1}_{3 \times 3} \) and is jointly optimized along with \( \varepsilon \) during the training phase to adjust the contribution of the learnable descriptive convolution.

Moreover, SSMs often introduce a larger number of hidden states to remember long-range dependencies, leading to significant channel redundancy when visualizing the activation results for different channels \cite{VMamba}. To enhance the expressive power of different channels, we integrate efficient channel attention (ECA \cite{ECA}) into DVSS. This allows SSMs to concentrate on learning diverse channel representations, with subsequent channel attention selecting critical channels to prevent redundancy. DVSS is formulated as:
\begin{equation}
\begin{aligned}
\bm{Z}^{l} &= \operatorname{ESSM}\left(\operatorname{LN}\left(\bm{F}_{D}^{l}\right)\right) + \bm{F}_{D}^{l}, \\
\bm{F}_{D}^{n+1} &= \operatorname{ECA}\left(\operatorname{LN}\left(\bm{Z}^{l}\right)\right) + \operatorname{LDC}\left(\bm{F}_{D}^{n}\right) + \bm{Z}^{l},
\vspace{-4em}
\end{aligned}
\end{equation}
where \(\bm{F}_{D}^{n}\) represents the feature map of the \(\mathit{n}\)-layer and \(\bm{F}_{D}^{n+1}\) is the feature of the input at the next level. \(\operatorname{ESSM}\left ( \cdot  \right )\), \(\operatorname{LDC}\left ( \cdot  \right )\) and \(\operatorname{ECA}\left ( \cdot  \right )\) denote the ESSM, LDC and ECA operations as shown in Figure. \ref{fig:FusionMamba}(c), Figure. \ref{fig:FusionMamba}(d) and Figure. \ref{fig:FusionMamba}(e), respectively.  

\subsection{Dynamic Feature Enhancement Module}
\begin{figure}[t]
  \centering
  \includegraphics[width=0.85\linewidth]{ 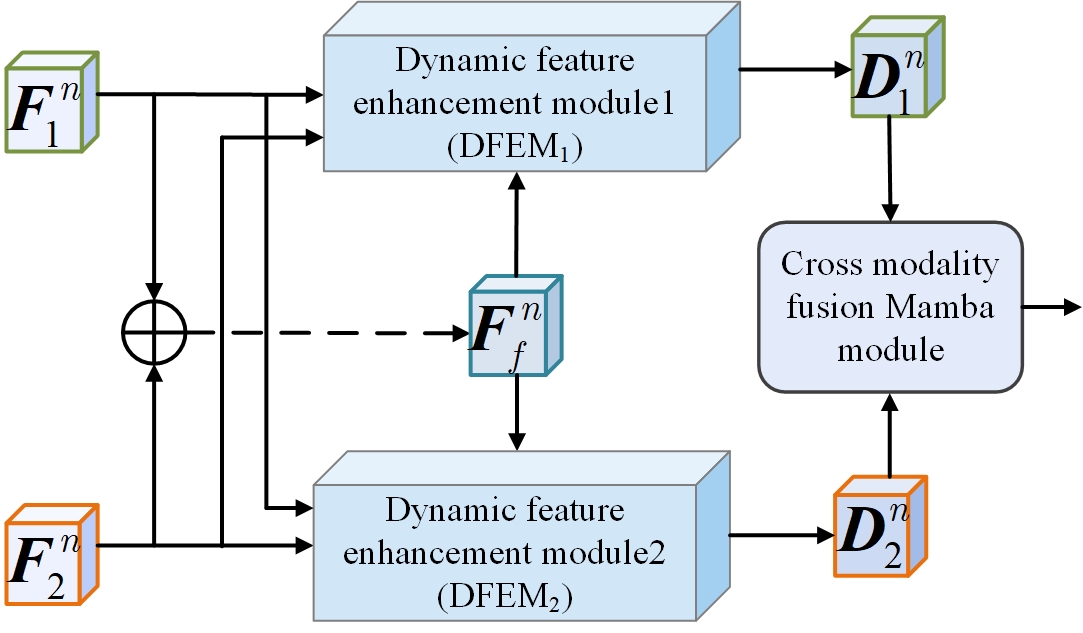}
  \vspace{-0.3em}
  \caption{Dynamic feature fusion module (DFFM). \(\bm{D}_{1}^{n}\) and \(\bm{D}_{2}^{n}\) are features; \(\bm{F}_{1}^{n}\) and \(\bm{F}_{2}^{n}\) are different modal features; \(\bm{F}_{f}^{n}\) is a coarse-grained feature fusion. \(\oplus\)is the element-wise addition operation.}
  \vspace{-2em}
  \label{fig:DFFM}
\end{figure}

The dynamic feature enhancement module is an important part of the designed fusion module (in Figure. \ref{fig:DFFM}). It aims to adaptively extract the texture detail features of different modalities through the dynamic feature enhancement mechanism and dynamically perceive the differences between different modalities. Here the features (\(\bm{D}_{1}^{n}\), \(\bm{D}_{2}^{n}\)) after passing the dynamic feature enhancement module (as shown in Fig. \ref{fig:DFEM}) are as follows:
\begin{equation}
\begin{aligned}
\bm{D}_{1}^{n} &= \bm{F}_{1}^{n} \oplus \bm{T}_{1}^{n} \oplus \delta\left(GAP\left(\bm{T}_{2}^{n} - \bm{T}_{1}^{n}\right)\right) \odot \bm{F}_{f}^{n}, \\
\bm{D}_{2}^{n} &= \bm{F}_{2}^{n} \oplus \bm{T}_{2}^{n} \oplus \delta\left(GAP\left(\bm{T}_{1}^{n} - \bm{T}_{2}^{n}\right)\right) \odot \bm{F}_{f}^{n},
\end{aligned}
\end{equation}
where the inputs are two different modal features (\(\bm{F}_{1}^{n}\), \(\bm{F}_{2}^{n}\)) as well as a coarse-grained feature fusion (\(\bm{F}_{f}^{n}\)), which is obtained by element-wise addition of different modal features. We apply the LDC module (see Section 3.2 for details) to the feature maps of each modality. The LDC module can amplify the texture information by adjusting the convolutional kernel weights, which enhances the model's ability to perceive the texture features, resulting in enhanced feature maps \(\bm{T}_{1}^{n}\) and \(\bm{T}_{2}^{n}\). These processed features of different modalities are then subtracted by elements to obtain difference feature maps. These maps are then passed through a global pooling (\(\mathit{GAP\left(\cdot\right)}\)) and a sigmoid function (\(\delta\)) operation to compute the difference weights between the output feature maps. These weights are applied to the fused feature maps (\(\bm{F}_{f}^{n}\)) and the original feature maps (\(\bm{F}_{1}^{n}\) or \(\bm{F}_{2}^{n}\)) in order to amplify the differences and efficiently extract and enhance the complementary features and texture details inherent in the images, thus improving the overall fusion performance.

\subsection{Cross-Modal Fusion Mamba Module}
\begin{figure}[t]
  \centering
  \includegraphics[width=0.82\linewidth]{ 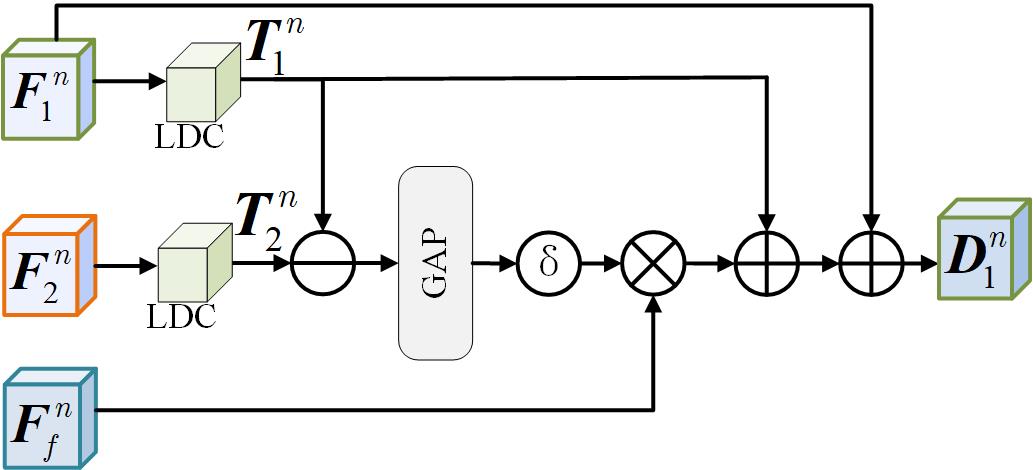}
  \caption{Dynamic feature enhancement module (\(\mathrm{DFEM}_{1})\). \(\bm{T}_{1}^{n}\) and \(\bm{T}_{2}^{n}\) are enhanced feature maps. These maps are then passed through  a global pooling (\(\mathit{GAP\left(\cdot\right)}\))) operation and a sigmoid function (\(\delta\)) to compute the difference weights between the output feature maps. \(\otimes\)  and \(\oplus\) are the element-wise multiplication and addition operations.}
  \vspace{-0.8em}
  \label{fig:DFEM}
\end{figure}

\begin{figure}[t]
  \centering
  \includegraphics[width=\linewidth]{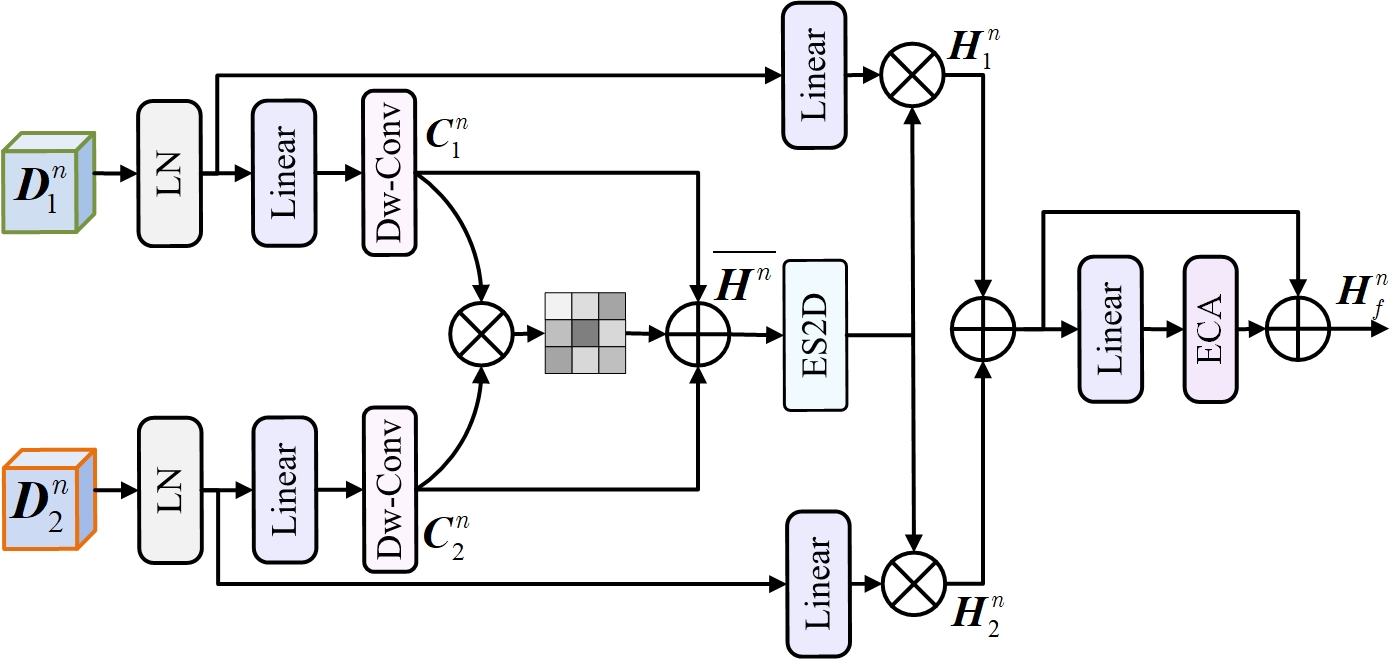}
   \vspace{-1.5em}
  \caption{Cross modal fusion mamba module (CMFM). Dw-Conv: depthwise convolution; ECA: effecient channel attention; ES2D: the efficient 2D scanning. \(\bm{C}_{1}^{n}\) and \(\bm{C}_{2}^{n}\) are the feature maps obtained after applying depthwise convolution to the inputs \(\bm{D}_{1}^{n}\) and \(\bm{D}_{2}^{n}\), respectively, used to extract finer spatial features. \(\bm{H}_{1}^{n}\) and \(\bm{H}_{2}^{n}\) are the hybrid features.
 }
  \vspace{-1.0em}
  \label{fig:CMFM}
\end{figure}

The output features from the DFEM, shown in Figure. \ref{fig:DFEM}, are further fed into the cross-modal fusion Mamba module (CMFM, shown in Figure. \ref{fig:CMFM}), which is used for fine-grained fusion and exploring the correlation of the information between different modalities. (\(\bm{D}_{1}^{n}\), \(\bm{D}_{2}^{n}\)) are first passed through depthwise convolution to obtain \(\bm{C}_{1}^{n}\) and \(\bm{C}_{2}^{n}\). Then, the hybrid features \(\overline{\bm{H}^{n}}\) are generated as follows:
\begin{equation}
\begin{aligned}
\bm{C}_{1}^{n} &= \mathrm{Dwc}(\mathrm{Linear}(\mathrm{LN}(\bm{D}_{1}^{n}))), \\
\bm{C}_{2}^{n} &= \mathrm{Dwc}(\mathrm{Linear}(\mathrm{LN}(\bm{D}_{2}^{n}))), \\
\overline{\bm{H}^{n}} &= \bm{C}_{1}^{n} \otimes \bm{C}_{2}^{n} \oplus \bm{C}_{1}^{n} \oplus \bm{C}_{2}^{n},
\end{aligned}
\end{equation}
where \(\mathrm{Dwc}\left ( \cdot  \right )\) is the depthwise convolution operation. \(\otimes\) and \(\oplus\) are the element-wise multiplication and addition operations, respectively.

The hybrid enhanced features are then fed into the efficient spatial scanning 2D (ES2D) \cite{EVMamba} layer to capture long-term spatial dependencies. Following this, the features pass through an efficient channel attention (ECA) operation to reduce channel redundancy, as outlined in the process below:
\begin{equation}
\begin{gathered}
\bm{H}_{1}^{n}= \mathrm{LN}\left ( \mathrm{ES2D}\left ( \left ( \overline{\bm{H}^{n}} \right ) \right ) \right )\otimes \left (\mathrm{Linear}\left ( \bm{D}_{1}^{n} \right )  \right ),\\
\bm{H}_{2}^{n}= \mathrm{LN}\left ( \mathrm{ES2D}\left ( \left ( \overline{\bm{H}^{n}} \right ) \right ) \right )\otimes \left (\mathrm{Linear}\left ( \bm{D}_{2}^{n} \right )  \right ),\\
\bm{H}_{f}^{n}=\mathrm{ECA}\left(\mathrm{Linear}\left ( \bm{H}_{1}^{n}\oplus\bm{H}_{2}^{n} \right ) \right )\oplus \left ( \bm{H}_{1}^{n}\oplus\bm{H}_{2}^{n} \right ).
\end{gathered}
\end{equation}

The final fused feature map of each layer (\(\bm{X}^{n}\)) is calculated using Eq. (10):

\begin{equation}
\begin{gathered}
\bm{X}^{n}=\bm{H}_{f}^{n}\oplus \bm{F}_{1}^{n}\oplus \bm{F}_{2}^{n},
\end{gathered}
\end{equation}
where \(n\) represents the \(n\)-th layer, and \(\bm{F}_{1}^{n}\) and \(\bm{F}_{2}^{n}\) denote the feature maps from different modalities input to this layer.
\subsection{Loss Function}
To ensure the extraction of meaningful information during training, we introduce three types of loss functions: the intensity loss \(\mathcal{L}_\text{int}\), the texture loss \(\mathcal{L}_\text{text}\), and the structure loss \(\mathcal{L}_\text{ssim}\). The total loss \(\mathcal{L}_\text{total}\) can be shown as follows:
\begin{equation}
  \mathcal{L}_\text{total} = \alpha_\text{1}\mathcal{L}_\text{int}+ \alpha_\text{2}\mathcal{L}_\text{text}+ \alpha_\text{3}\mathcal{L}_\text{ssim},
\end{equation}
where \(\alpha_\text{1}\), \(\alpha_\text{2}\), and \(\alpha_\text{3}\) are the weights to control the trade-off between \(\mathcal{L}_\text{int}\), \(\mathcal{L}_\text{text}\), and \(\mathcal{L}_\text{ssim}\).

Integrating more texture details is essential to improve visual quality. We use gradient loss to ensure that more fine-grained details are preserved, which is defined as
\begin{equation}
  \mathcal{L}_\text{text} = \frac{1}{HW}\left\|\triangledown \bm{I}_{f}-\text{max}\left ( \triangledown \left|\bm{I}_\text{1} \right|,\triangledown \left|\bm{I}_\text{2} \right|\right ) \right\|_\text{1},
\end{equation}
where \(H\) and \(W\) are the height and width of the image, respectively. \(\bm{I}_{1}\) and \(\bm{I}_{2}\) denote the different source images. \(\bm{I}_{f}\) denotes the fused image. \(\text{max}()\) denotes the max operation. \(\nabla \bm{I}\) denotes the gradient of the image \(\bm{I}\), capturing edge information in both horizontal and vertical directions. 

Generally, there is significant intensity information and contrast information in source images. We employ intensity loss to ensure that appropriate intensity information is retained. It can be defined as
\begin{equation}
  \mathcal{L}_\text{int} = \frac{1}{HW}\left\|\bm{I}_{f}-\text{max}(\bm{I}_\text{1},\bm{I}_\text{2} ) \right\|_\text{2}.
\end{equation}

SSIM \cite{SSIM} can measure the degree of distortion and the degree of similarity between two images. We use SSIM loss to guarantee structure similarity between the fused image and source images. It can be defined as
\begin{equation}
  \mathcal{L}_\text{ssim} = \frac{1}{2}(1 - \text{SSIM}(\bm{I}_\text{1}, \bm{I}_{f})) + \frac{1}{2}(1 - \text{SSIM}(\bm{I}_\text{2}, \bm{I}_{f})).
\end{equation}
\section{Experiment}

\subsection{Setup}

\subsubsection{Datasets} 
For the IVF fusion task, we choose the KAIST \cite{KAIST} dataset to train our FusionMamba.The KAIST dataset captures images from a variety of scenes (including schoolyards, streets, and a variety of regular traffic scenes in the countryside) during the daytime and nighttime, respectively, and each image is available in two versions, infrared vs. visible light images. This dataset has a rich variety of scenes and pairs of data, so we selected 70,000 pairs of infrared and visible light images from this dataset for training. These images were also converted to grayscale and resized to 256 × 256. The test dataset consists of the MSRS test set (361 pairs), the RoadScene test set (61 pairs), and the TNO \cite{TNO} (21 pairs). These datasets contain a variety of images taken during daytime and nighttime, and most of the images are taken on roads, including objects such as people, cars, bicycles, and road signs, so that the fusion performance can be comprehensively verified.

For MIF tasks, we use Harvard public medical dataset \footnote{http://www.med.harvard.edu/AANLIB/home.html} as the training and test dataset independently. There are 166 CT-MRI image pairs, 329 PET-MRI image pairs and 539 SPECT-MRI image pairs. The size of images is 256×256. We augment image pairs to 30000 via image rotation, which can improve the model and help to combat potential overfitting. 21 pairs of test images are randomly chosen to evaluate the model.

The GFP dataset released by John Innes Centre \cite{GFPPC} is employed in our experiments.  The database contains 148 pairs of pre-registered GFP and PC images of Arabidopsis thealiana cells with the same size of 358 × 358 pixels. We also augment image pairs to 30000 via image rotation.  20 pairs of test images are randomly chosen to evaluate the model.

\subsubsection{Implementation Details and Metrics}
The batch size is 2 and Adam optimizer with a learning rate 0.0002 is used. \(\alpha_\text{1}\), \(\alpha_\text{2}\), and \(\alpha_\text{3}\) are set as 100, 10, 1, respectively. The experiments are conducted using Nvidia GeForce RTX 3090 GPU and 3.60 GHz Intel Core i9-9900K CPU with Pytorch.

Fusion performance is evaluated using six key metrics \cite{Q,SSIM}:  structure content difference (SCD), multiscale structural similarity index measure (MS-SSIM), gradient-based metric (\(Q^{AB/F}\) ), feature mutual information (FMI), and visual information fidelity (VIF) \cite{VIF}. SCD evaluates the structural and content differences between the fused image and the input images. MS-SSIM evaluates structural similarity across scales, \(Q^{AB/F}\)  assesses edge information retention, FMI measures feature information preservation, and VIF calculates visual fidelity comprehensively. These metrics together offer a comprehensive evaluation of fusion performance across different criteria.

\begin{figure*}[ht]
  \centering
  \includegraphics[width=\textwidth]{ 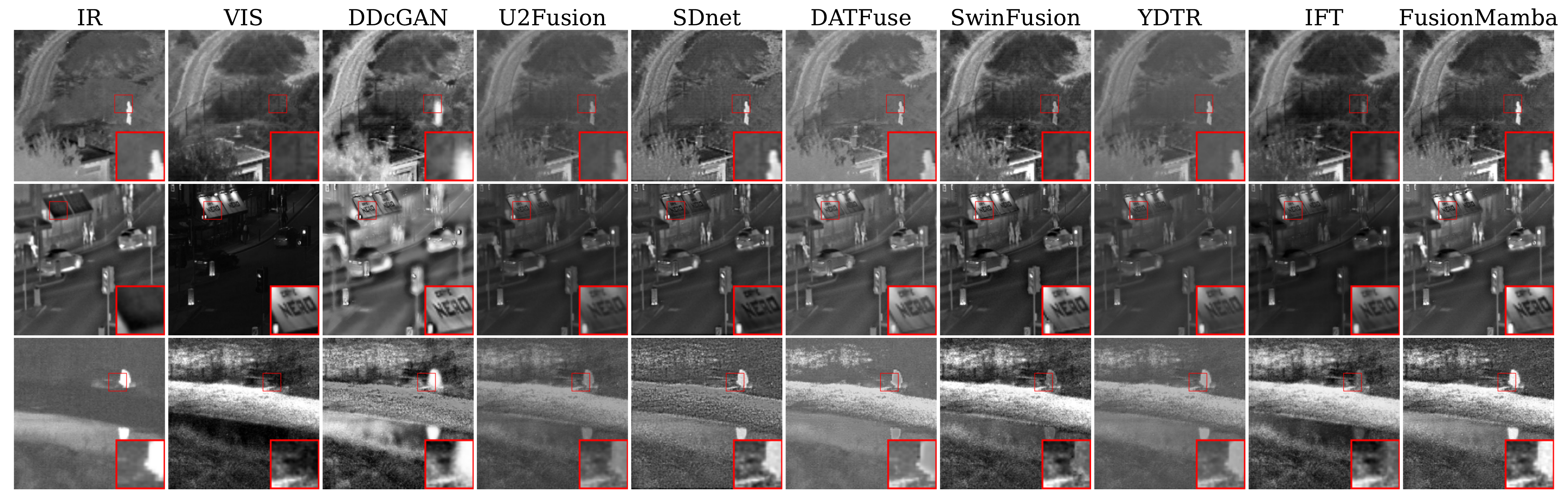}
  \vspace{-1.0em}
  \caption{Qualitative results on fusion of three typical IR and VIS images on the TNO \cite{TNO} dataset. Comparison methods include DDcGAN \cite{DDCGAN}, U2Fusion \cite{U2Fusion}, SDnet \cite{SDNet}, DATFuse \cite{DATFuse}, SwinFusion \cite{SwinFusion}, YDTR \cite{YDTR}, and IFT \cite{IFT}.}
  \vspace{-1.0em}
  \label{fig:TNO}
\end{figure*}

\begin{figure*}[ht]
  \centering
  \vspace{-0.6em}
  \includegraphics[width=\textwidth]{ 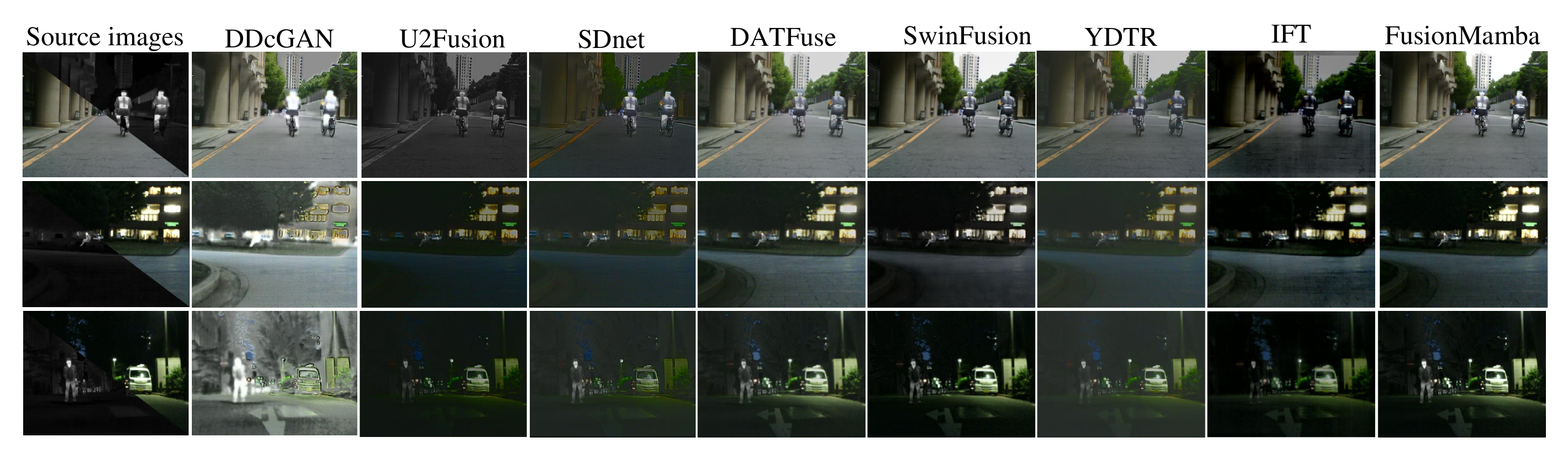}
  \vspace{-1.8em}
  \caption{Qualitative results on fusion of three typical IR and VIS images on the MSRS \cite{MSRS} dataset.}
  \vspace{-1.0em}
  \label{fig:MSRS}
\end{figure*}

\begin{figure*}[ht]
  \centering
  \vspace{-0.6em}
  \includegraphics[width=\textwidth]{ 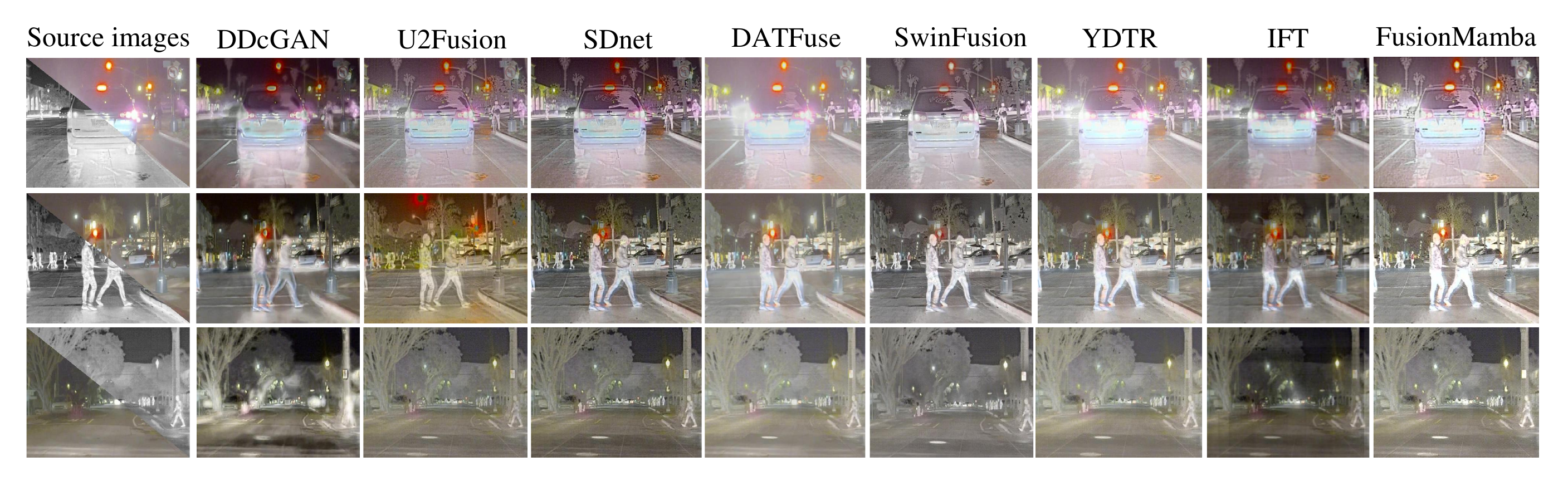}
  \vspace{-1.8em}
  \caption{Qualitative results on fusion of three typical IR and VIS images on the RoadScene \cite{RoadScene} dataset.}
  \vspace{-1.0em}
  \label{fig:RoadScene}
\end{figure*}

\subsubsection{Comparison Approaches}
We perform extensive comparison experiments with state-of-the-art methods to demonstrate the validity of FusionMamba. These comparison methods include CSMCA \cite{Me2}, U2Fusion \cite{U2Fusion}, DDcGAN \cite{DDCGAN}, SDnet \cite{SDNet}, DATFuse \cite{DATFuse}, YDTR \cite{YDTR}, MATR \cite{MATR}, IFT \cite{IFT} and SwinFusion \cite{SwinFusion}. CSMCA \cite{Me2} is the traditional fusion method. DATFuse \cite{DATFuse}, DDcGAN \cite{DDCGAN}, and U2Fusion \cite{U2Fusion} are CNN-based fusion methods. MATR \cite{MATR}, YDTR \cite{YDTR}, IFT and SwinFusion \cite{SwinFusion} are Transformer-based structures. We use the public codes with corresponding parameter settings.

\subsection{Infrared and Visible Image Fusion}
As illustrated in Figures \ref{fig:TNO}, \ref{fig:MSRS}, and \ref{fig:RoadScene}, we provide a detailed qualitative comparison of our method against several state-of-the-art algorithms across multiple infrared-visible fusion datasets, including TNO \cite{TNO}, MSRS \cite{MSRS}, and RoadScene \cite{RoadScene}. Across all datasets, our method consistently demonstrates superior performance, particularly excelling at preserving critical target details and texture information. In contrast, networks such as SDNet \cite{SDNet} and DATFuse \cite{DATFuse} struggle to accurately capture scene information from visible light images due to insufficient global information interaction and poor intensity control. Even methods such as YDTR \cite{YDTR}, IFT \cite{IFT}, and SwinFusion \cite{SwinFusion}, which retain some texture details, still suffer from thermal radiation contamination, reducing the clarity of infrared targets.

Our method effectively addresses these challenges by mitigating issues such as thermal target degradation, spectral contamination, and texture blurring. The visual results across different datasets highlight FusionMamba's strong ability to fully utilize and integrate complementary and shared information from multiple image modalities, overcoming the limitations of existing fusion networks. This advantage is based on two key innovations. First, the dynamically enhanced FusionMamba module excels at leveraging inter-modal feature links, enhancing complementary information, and revealing critical shared features. Second, the efficient Mamba framework provides strong global context awareness, which improves feature extraction and reconstruction. 

The quantitative analysis in Tables \ref{tab:TNO}, \ref{tab:MSRS}, and \ref{tab:RoadScene} further confirms the superior performance of FusionMamba. Specifically, FusionMamba achieves the highest score in the $Q^{AB/F}$ metric, demonstrating its exceptional edge-preservation capabilities. It also significantly outperforms competing methods in VIF, indicating its ability to produce visually appealing fused images with minimal perceptual distortion. Additionally, the FMI and MS-SSIM metrics show that FusionMamba effectively maintains both texture and structural similarity across modalities. Finally, the SCD metric confirms that our method minimizes texture distortion better than other approaches. Both qualitative and quantitative results demonstrate that FusionMamba not only excels at maintaining visual clarity, but also achieves a well-rounded fusion performance across key evaluation metrics, establishing itself as a leading method in image fusion.

\begin{table}[ht]
\tiny 
\caption{Quantitative analysis results for the infrared and visible (IR-VIS) (TNO \cite{TNO}) task. The best results are highlighted in bold. VIF: visual information fidelity; SCD: structure content difference; MS-SSIM: multiscale structural aimilarity index measure; \(Q^{AB/F}\): gradient-based metric; FMI: feature mutual information.}
\label{tab:TNO}
\setlength{\tabcolsep}{0.5pt} 
\renewcommand{\arraystretch}{1.2} 
\begin{tabularx}{\linewidth}{X*{6}{>{\centering\arraybackslash}X}}
\toprule
Method & VIF $\uparrow$ & SCD $\uparrow$ & $Q^{AB/F} \uparrow$ & MS-SSIM $\uparrow$& FMI $\uparrow$\\
\midrule
DDcG \cite{DDCGAN} &0.4278 &1.5734 &0.3608 &0.7241 &0.8592 \\
U2F \cite{U2Fusion} & 0.4736& 1.6390& 0.3539& 0.8343& 0.8836 \\
SDnet \cite{SDNet}&0.4715 &1.6237 &0.4430 &0.8436 &0.8785  \\
DATF \cite{DATFuse}& 0.6052& 1.5297& 0.5071& 0.7595& 0.8730 \\
SwinF \cite{SwinFusion}& 0.6687& 1.8095& 0.5611& 0.9236& 08894 \\
YDTR \cite{YDTR}&0.5279 &1.6137 &0.4078 &0.8007 &0.8871 \\
IFT \cite{IFT}&0.6350 & 1.7817& 0.4668 &0.8606 &0.8853 \\
\textbf{Ours} & \textbf{0.7717}& \textbf{1.8314}& \textbf{0.5468}& \textbf{0.9331}& \textbf{0.8858}\\
\bottomrule
\end{tabularx}
\vspace{-1.5em}
\end{table}
\begin{table}[ht]
\tiny 
\vspace{-0.5em}
\caption{Quantitative analysis results for the IR-VIS (MSRS \cite{MSRS}) task.}
\label{tab:MSRS}
\setlength{\tabcolsep}{1pt} 
\renewcommand{\arraystretch}{1.2} 
\begin{tabularx}{\linewidth}{X*{5}{>{\centering\arraybackslash}X}}
\toprule
Method & VIF $\uparrow$ & SCD $\uparrow$ & $Q^{AB/F} \uparrow$ & MS-SSIM $\uparrow$& FMI $\uparrow$\\
\midrule
DDcG \cite{DDCGAN} &0.5629 &1.1473 &0.3595 &0.5157 &0.8909 \\
U2F \cite{U2Fusion}&0.4864 &1.0642 &0.3384 &0.7420 &0.9127 \\
SDnet\cite{SDNet}  &0.4991  &0.9863  &0.3766  &0.7253  &0.9097  \\
DATF \cite{DATFuse}&0.6688 &1.3691 &0.3789 &0.7150 &0.9043 \\
SwinF \cite{SwinFusion}&0.6903 &1.5952 &0.4223 &0.7317 &0.9086 \\
YDTR  \cite{YDTR}&0.5781 &1.1381 &0.3489 &0.7292 &0.9172 \\
IFT \cite{IFT}&0.5154 &1.1922 &0.2378 &0.6874 &0.9033\\
\textbf{Ours} &\textbf{0.8579} &\textbf{1.6866} &\textbf{0.5979} &\textbf{0.7361} &\textbf{0.9269}       \\
\bottomrule
\end{tabularx}
\vspace{-1.1em}
\end{table}

\begin{table}[ht]
\tiny 
\caption{Quantitative analysis results for the IR-VIS (RoadScene \cite{RoadScene}) task.}
\label{tab:RoadScene}
\setlength{\tabcolsep}{1pt} 
\renewcommand{\arraystretch}{1.2} 
\begin{tabularx}{\linewidth}{X*{5}{>{\centering\arraybackslash}X}}
\toprule
Method & VIF $\uparrow$ & SCD $\uparrow$ & $Q^{AB/F} \uparrow$ & MS-SSIM $\uparrow$ & FMI $\uparrow$ \\
\midrule
DDcG\cite{DDCGAN} & 0.4195 & 1.4078 & 0.2812 & 0.6105 & 0.8594 \\
U2F\cite{U2Fusion} & 0.5318 & 1.0740 & \textbf{0.5318} & 0.7045 & 0.8558 \\
SDnet \cite{SDNet} &0.6132 &1.2077 &0.5458 &0.7037 &0.8629\\
DATF\cite{DATFuse} & 0.5761 & 0.8498 & 0.3871 & 0.6714 & 0.8543 \\
SwinF\cite{SwinFusion} & 0.4886 & 1.4657 & 0.4157 & 0.6822 & 0.8436 \\
YDTR\cite{YDTR} & 0.5817 & 1.2114 & 0.4565 & 0.7147 & 0.8621 \\
IFT\cite{IFT} & 0.4682 & 1.2833 & 0.2914 & 0.6865 & 0.8568 \\
\textbf{Ours} &\textbf{0.5722}  &\textbf{1.5810}  &0.4321  &\textbf{0.6739}  &\textbf{0.9162}  \\
\bottomrule
\end{tabularx}
\vspace{-1.0em}
\end{table}

\subsection{Multimodal Medical Image Fusion}
\begin{figure*}[ht]
  \centering
  \includegraphics[width=\textwidth]{ 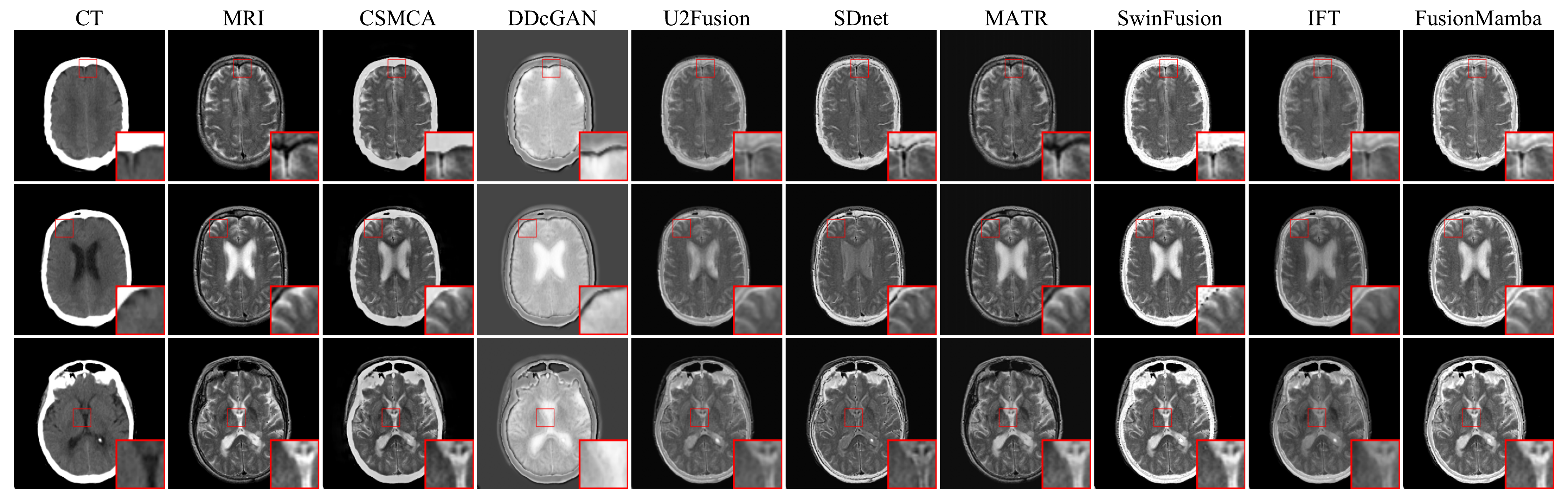}
  \vspace{-1.5em}
  \caption{Qualitative results of the CT and MRI image fusion task on the Harvard public medical dataset.}
  \vspace{-1.0em}
  \label{fig:CTMRI}
\end{figure*}

We present three typical medical image fusion tasks: CT-MRI image fusion, PET-MRI image fusion, and SPECT-MRI image fusion tasks. For the qualitative analysis, Figure. \ref{fig:CTMRI} displays the CT-MRI fusion task, including CT images, MRI images, and the fusion results of CSMCA \cite{Me2}, U2Fusion \cite{U2Fusion}, FusionGAN \cite{FusionGAN}, SDNet \cite{SDNet}, MATR \cite{MATR}, SwinFusion \cite{SwinFusion}, IFT \cite{IFT}, and the proposed FusionMamba. We zoomed in on the area in the red box, demonstrating that our fusion images can simultaneously retain the dense information of CT and the texture details of MRI. In Figure. \ref{fig:CTMRI}, FusionGAN and U2Fusion \cite{U2Fusion} exhibit unsatisfactory brightness and clarity, with some gray matter blurring the texture details. Particularly, FusionGAN shows artifacts and redundant information. MATR \cite{MATR}preserves structural details but weakens dense structures. SDNet \cite{SDNet} and CSMCA \cite{Me2} retain dense structures but lose some edge details. SwinFusion \cite{SwinFusion} achieves satisfactory fusion but with excessively sharp edges. In contrast, our FusionMamba retains more texture details while maintaining appropriate dense structures. Visually, they appear more natural with enhanced contrast.

\begin{figure*}[ht]
  \centering
  \includegraphics[width=\textwidth]{ 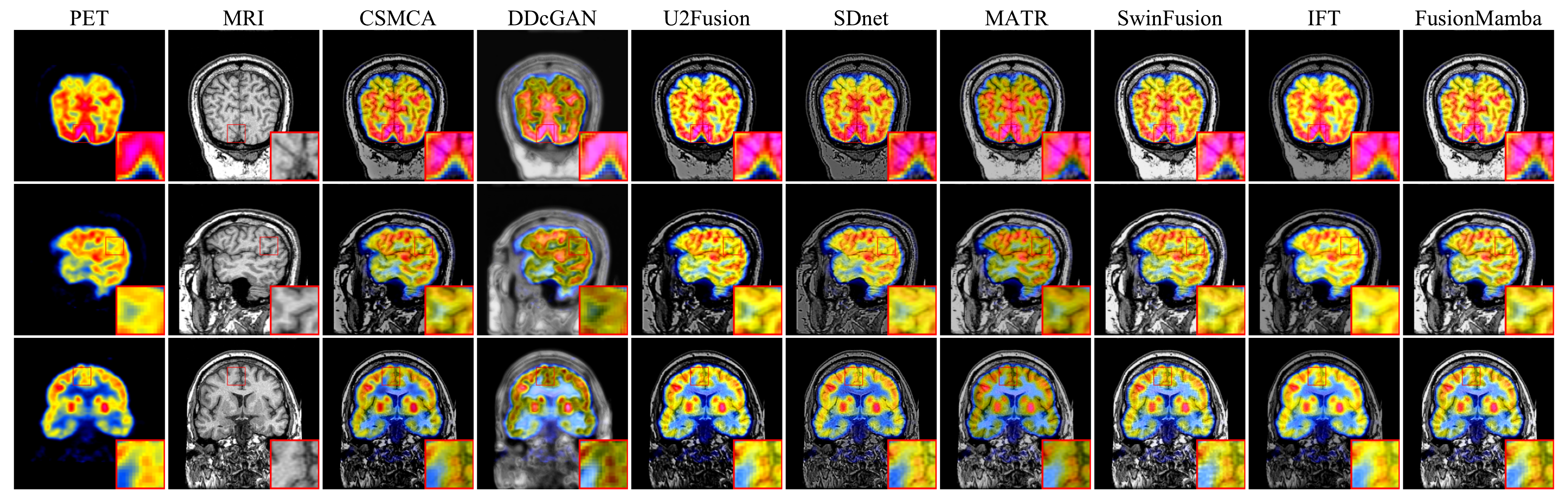}
  \vspace{-1.5em}
  \caption{Qualitative results of the PET and MRI image fusion task on the Harvard public medical dataset.}
  \vspace{-0.5em}
  \label{fig:PETMRI}
\end{figure*}

\begin{figure*}[ht]
  \centering
  \includegraphics[width=\textwidth]{ 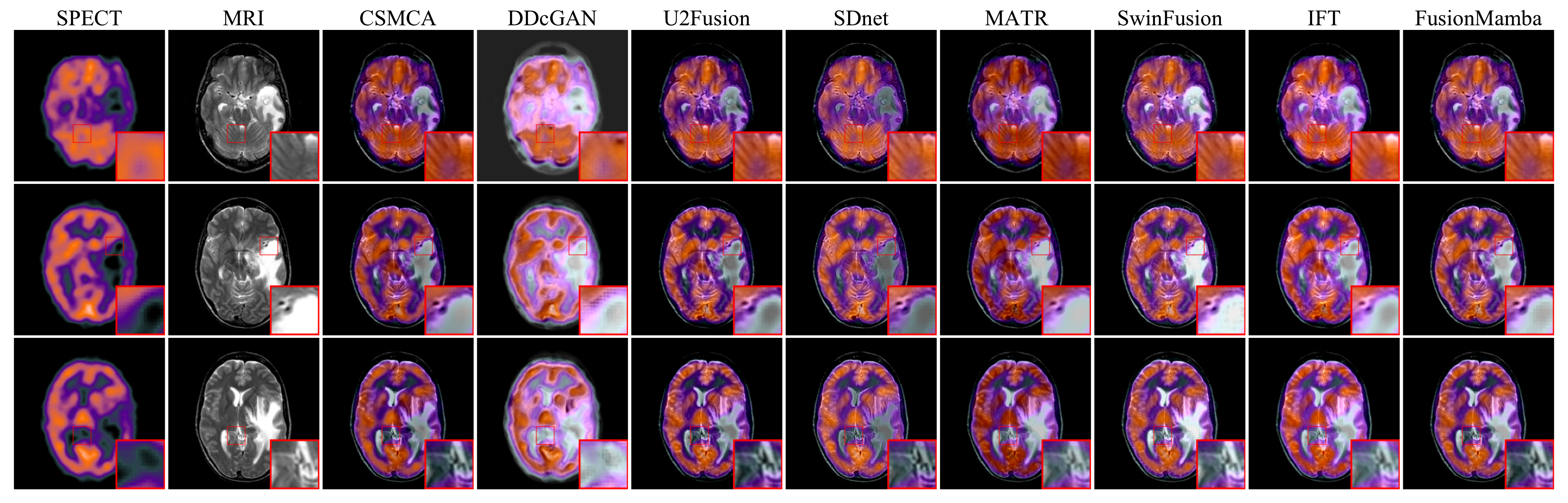}
  \vspace{-1.5em}
  \caption{Qualitative results of the SPECT and MRI image fusion image task on the Harvard public medical dataset.}
  \vspace{-1.0em}
  \label{fig:SPECTMRI}
\end{figure*}

Figure. \ref{fig:PETMRI} and Figure. \ref{fig:SPECTMRI} present the PET-MRI fusion task and the SPECT-MRI fusion task. FusionGAN's fusion images have lower contrast and missing texture details, while CSMCA \cite{Me2} and IFT \cite{IFT} preserve texture information well. However, the darker color indicates insufficient preservation of functional information. Additionally, IFT \cite{IFT}, U2Fusion \cite{U2Fusion}, and SDNet \cite{SDNet} retain functional features but with less sharp texture details in salient regions. SwinFusion \cite{SwinFusion} shows color distortion due to over-sharpening. In contrast, our FusionMamba preserves clear edges and texture details, with a color distribution more similar to PET images, enhancing visual perception. focusing on the SPECT-MRI fusion task, similar to PET-MRI fusion, our FusionMamba excels in capturing more details and retaining appropriate color information from SPECT images.

Table \ref{tab:CTMRI} displays the quantitative comparison results for six metrics in the CT-MRI fusion task. FusionMamba achieves optimal results (average values) in VIF, SCD, \(Q^{AB/F}\), MS-SSIM, and FMI, indicating higher structure similarity, enhanced contrast, and better visual effects. Similar trends are observed in Table \ref{tab:PETMRI} for the PET-MRI fusion task and Table \ref{tab:SPECTMRI} for the SPECT-MRI fusion task, where FusionMamba consistently outperforms other methods across various metrics, demonstrating superior fusion performance in retaining functional and morphological information.

\begin{table}[ht]
\vspace{-2em}
\tiny 
\caption{Comparison results in the CT-MRI Task on the Harvard public medical dataset.}
\label{tab:CTMRI}
\setlength{\tabcolsep}{0.5pt} 
\renewcommand{\arraystretch}{1.2} 
\begin{tabularx}{\linewidth}{X*{6}{>{\centering\arraybackslash}X}}
\toprule
Method & VIF $\uparrow$ & SCD $\uparrow$ & $Q^{AB/F} \uparrow$ & MS-SSIM $\uparrow$& FMI $\uparrow$\\
\midrule
CSMCA \cite{Me2} & 0.4745 & 1.1372 & 0.5986& 0.9444 & \textbf{0.8800} \\
DDcG \cite{DDCGAN}&0.1709 &0.6229 &0.2787 &0.5774 &0.8518 \\
U2F \cite{U2Fusion} & 0.4246 & 0.8399 & 0.2865  & 0.8355 & 0.8619 \\
SDnet \cite{SDNet} & 0.4224 & 0.9895 & 0.4797  & 0.8672 & 0.8350 \\
MATR \cite{MATR} & 0.4556 & 0.3468 & 0.5044  & 0.6095 & 0.8600 \\
SwinF \cite{SwinFusion} & \textbf{0.6075} & 1.2659 & 0.6312  & 0.9042 & 0.8339 \\
IFT \cite{IFT} & 0.5016 & 1.1188 & 0.4571  & 0.8928 & 0.8687 \\
\textbf{Ours} & 0.5750 & \textbf{1.5884} & \textbf{0.6429}  & \textbf{0.9462} & 0.8624 \\
\bottomrule
\end{tabularx}
\vspace{-3.0em}
\end{table}

\begin{table}[ht]
\vspace{-2em}
\tiny 
\caption{Comparison results in the PET-MRI Task on Harvard public medical dataset.}
\label{tab:PETMRI}
\setlength{\tabcolsep}{0.5pt} 
\renewcommand{\arraystretch}{1.2} 
\begin{tabularx}{\linewidth}{X*{6}{>{\centering\arraybackslash}X}}
\toprule
Method & VIF $\uparrow$ & SCD $\uparrow$ & $Q^{AB/F} \uparrow$ & MS-SSIM $\uparrow$ & FMI $\uparrow$\\
\midrule
CSMCA \cite{Me2} & 0.5515& 0.7892& 0.6505& 0.9156& 0.8401 \\
DDcG \cite{DDCGAN}&0.1671 &0.1535 &0.1330 &0.5349 &0.7824 \\
U2F \cite{U2Fusion} & 0.4618& 1.2809& 0.5153& 0.9074& 0.8232 \\
SDnet \cite{SDNet}& 0.5061& 1.0270& 0.5831& 0.8971& 0.8208 \\
MATR \cite{MATR}& 0.6033& 0.6442& 0.7254& 0.8290& 0.8377 \\
SwinF \cite{SwinFusion}& 0.6775& 1.3148& 0.7254& 0.9261& \textbf{0.8650} \\
 IFT \cite{IFT}& 0.6149& \textbf{1.4068}& 0.5501& 0.9129&0.8431 \\
\textbf{Ours} & \textbf{0.6920}& 1.3906& \textbf{0.7406}& \textbf{0.9362}& 0.8616\\
\bottomrule
\end{tabularx}
\vspace{-3.0em}
\end{table}

\begin{table}[ht]
\vspace{-2em}
\tiny 
\caption{Comparison results in the SPECT-MRI Task on Harvard public medical dataset.}
\label{tab:SPECTMRI}
\setlength{\tabcolsep}{0.5pt} 
\renewcommand{\arraystretch}{1.2} 
\begin{tabularx}{\linewidth}{X*{6}{>{\centering\arraybackslash}X}}
\toprule
Method & VIF $\uparrow$ & SCD $\uparrow$ & $Q^{AB/F} \uparrow$ & MS-SSIM $\uparrow$& FMI $\uparrow$\\
\midrule
CSMCA \cite{Me2} & 0.6051& 0.1854& 0.6301& 0.9144& 0.8827 \\
DDcG \cite{DDCGAN}&0.2106 &0.6634 &0.1487 &0.5824 &0.8268 \\
U2F \cite{U2Fusion} & 0.4525& 0.5633& 0.4789& 0.9209& 0.8600 \\
SDnet \cite{SDNet}& 0.4715& 1.1937& 0.4430& 0.8437& 0.8785 \\
MATR \cite{MATR}& 0.7235& 0.1706& 0.6620& 0.8883& 0.8880 \\
SwinF \cite{SwinFusion}& 0.6147& 1.1665& 0.6468& \textbf{0.9568}& 0.8887 \\
IFT \cite{IFT}& 0.6024& 1.1504& 0.6127& 0.9196&0.8770 \\
\textbf{Ours} & \textbf{0.7891}& \textbf{1.2106}& \textbf{0.7080}& 0.9545& \textbf{0.8901} \\
\bottomrule
\end{tabularx}
\vspace{-3.5em}
\end{table}

\subsection{Multimodal Biomedical Image Fusion}
\begin{figure*}[ht]
  \centering
  \includegraphics[width=\textwidth]{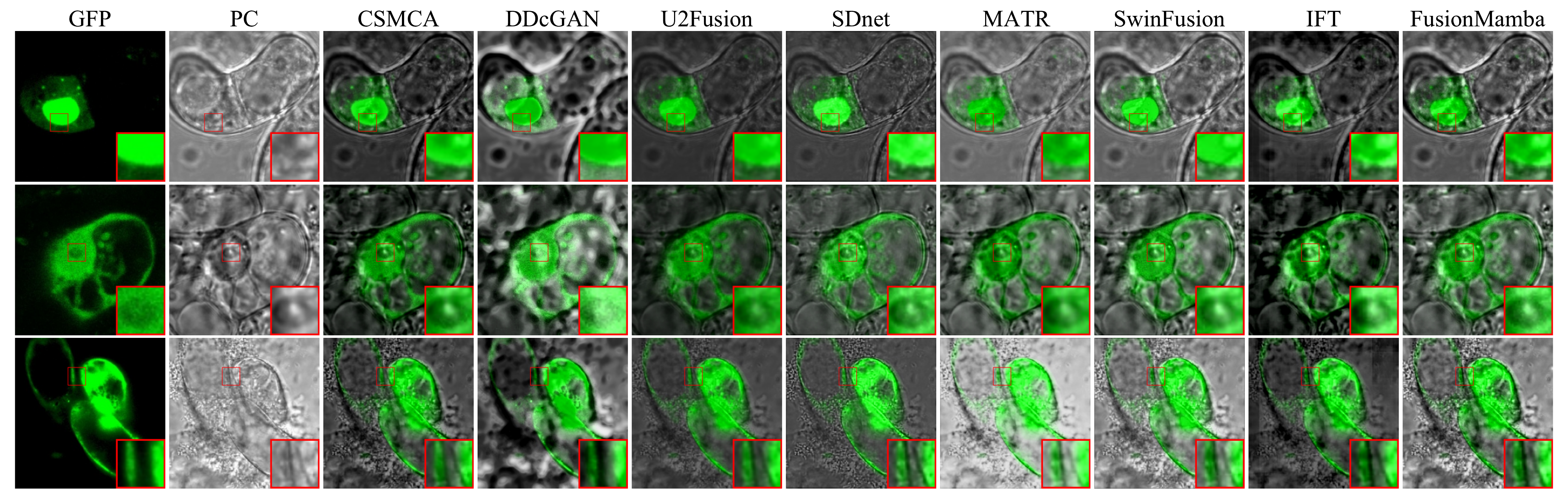}
  \vspace{-1.5em}
  \caption{ Qualitative results on fusion of three typical GFP and PC images of Arabidopsis thealiana cells on John Innes Centre \cite{GFPPC}.}
  \label{fig:GFPPC}
  \vspace{-1.0em}
\end{figure*}

To demonstrate the generalization capability of FusionMamba, we conduct green fluorescent protein (GFP) and phase contrast (PC) image fusion. GFP images provide functional information related to protein distribution, while PC images contain rich details of cell structure including nucleus and mitochondria. GFP and PC fusion images can promote biological research such as gene expression and protein function analysis. In Figure. \ref{fig:GFPPC} and Table \ref{tab:GFPPC}, it can be found that CSMCA \cite{Me2}, DDcGAN \cite{DDCGAN}, U2Fusion \cite{U2Fusion} and IFT \cite{IFT} can well preserve color information, but there is a certain loss of texture details. U2Fusion \cite{U2Fusion} and MATR \cite{MATR} can well retain texture details, but there is slight color distortion. It is worth noting that both SwinFusion \cite{SwinFusion} and FusionMamba can effectively preserve texture and color information, which proves the effectiveness of cross fusion strategies in image fusion.

\begin{table}[ht]
\tiny 
\caption{Comparison results in the GFP-PC task.}
\label{tab:GFPPC}
\setlength{\tabcolsep}{0.5pt} 
\renewcommand{\arraystretch}{1.2} 
\begin{tabularx}{\linewidth}{X*{6}{>{\centering\arraybackslash}X}}
\toprule
Method & VIF $\uparrow$ & SCD $\uparrow$ & $Q^{AB/F}$ $\uparrow$ & MS-SSIM $\uparrow$ & FMI $\uparrow$ \\
\midrule
CSMCA \cite{Me2} & 0.7609 & 1.4303 & 0.6475 & 0.9453 & 0.8723 \\
DDcG \cite{DDCGAN} & 0.3360 &1.2724 & 0.2261 & 0.6643 & 0.7836 \\
U2F \cite{U2Fusion} & 0.6085 & 1.3442 & 0.3329 & 0.8562 & 0.8616 \\
SDnet \cite{SDNet} & 0.5046 & 1.3040 & 0.49834 & 0.9023 & 0.8405 \\
MATR \cite{MATR} & 0.8147 & 0.8400 & \textbf{0.7205} & 0.8893 & 0.8637 \\
SwinF \cite{SwinFusion} & 0.8159 & 1.2185 & 0.6727 & 0.9138 & 0.8612 \\
IFT \cite{IFT} & 0.5652 & 1.3588 & 0.4659 & 0.8993 & 0.8368 \\
\textbf{Ours} & \textbf{0.8503} & \textbf{1.4623} & 0.6863 & \textbf{0.9328} & \textbf{0.8705} \\
\bottomrule
\end{tabularx}
\vspace{-1em}
\end{table}

\begin{table*}[t]
\centering
\caption{Computational cost analysis. FLOPs: floating point operations per second.}
\tiny
\vspace{-0.5em}
\label{tab:cost}
\setlength{\tabcolsep}{1pt} 
\renewcommand{\arraystretch}{1.2} 
\begin{tabularx}{\linewidth}{X*{7}{>{\centering\arraybackslash}X}}
\toprule
Method &CSMCA \cite{Me2} & U2F \cite{U2Fusion} & DDcG \cite{DDCGAN} &  IFCNN \cite{IFCNN} & IFT \cite{IFT} & SwinF \cite{SwinFusion} & Ours \\
\midrule
Runtime(s) & 69 & 0.20 &0.82 & 0.19 & 0.36 & 0.23 & \textbf{0.13} \\
Flops (G) & -- & 345.75 &211.60 & 77.94 & 184.56 & 135.17 & \textbf{26.48} \\
\bottomrule
\end{tabularx}
\vspace{-1.0em}
\end{table*}

\subsection{Computational Cost Analysis}
The complexity evaluation in Table \ref{tab:cost} assesses the operational efficiency of various methods by measuring floating-point operations per second (FLOPs) and running time. Specifically, the first image in the dataset is tested in an infrared and visible light fusion scenario to compute the FLOPs of each method. To better compare deep learning-based methods with traditional methods, we include a comparison with the advanced traditional algorithm CSMCA \cite{Me2} (implemented in Matlab). As shown in Table \ref{tab:cost}, we can observe that deep learning methods have a significant advantage in running time due to GPU acceleration. We compare the CNN-based U2Fusion \cite{U2Fusion} and IFCNN \cite{IFCNN} methods with the Transformer-based IFT \cite{IFT} and SwinFusion \cite{SwinFusion} methods. The results reveal that the Mamba-based approach has a notable advantage in terms of runtime, with lower FLOPs and average running time compared to CNN and Transformer methods. Importantly, our method demonstrates superior fusion performance when compared to mainstream image fusion algorithms.

\begin{table*}[t]
\caption{Ablation of network structure on the TNO \cite{TNO} dataset. LDC: learnable descriptive convolution; DVSS:  dynamic visual state space; EVSS: efficient visual state space block; DFFM: dynamic feature fusion module; DFEM: dynamic feature enhancement module; CMFM: cross-modal fusion mamba module.}
\tiny
\vspace{-0.8em}
\label{tab:ablation1}
\setlength{\tabcolsep}{1pt} 
\renewcommand{\arraystretch}{1.2} 
\begin{tabularx}{\linewidth}{>{\raggedright\arraybackslash}X*{6}{>{\centering\arraybackslash}X}}
\toprule
Method & VIF $\uparrow$ & SCD $\uparrow$ & $Q^{AB/F} \uparrow$ &  MS-SSIM $\uparrow$ & FMI $\uparrow$ \\
\midrule
w/o LDC &0.7461 &1.8247 &0.5283 &0.9210 &0.8799\\
w/o ECA &0.7543 &1.8253 &0.5411 &0.9298 &0.8810\\
DVSS$\to$EVSS & 0.6809 & 1.7917 & 0.5086 & 0.9006 & 0.8625 \\
DVSS$\to$Transformer & 0.6701 & 1.7329 & 0.5014  & 0.8812 & 0.8334 \\
DFFM$\to$DVSS & 0.7334 & 1.7881 & 0.5220  & 0.8990 & 0.8417 \\
w/o DFEM & 0.7459 & 1.7412 & 0.5129  & 0.8774 & 0.8719 \\
w/o CMFM & 0.7016 & 1.7615 & 0.5231  & 0.8895 & 0.8751 \\
Ours & \textbf{0.7717} & \textbf{1.8314} & \textbf{0.5468}  & \textbf{0.9331} & \textbf{0.8858} \\
\bottomrule
\end{tabularx}
\vspace{-1.8em}
\end{table*}

\begin{table}[t]
\caption{Ablation of loss function on the TNO \cite{TNO} dataset.}
\tiny
\label{tab:ablation2}
\setlength{\tabcolsep}{1pt} 
\renewcommand{\arraystretch}{1.2} 
\begin{tabularx}{\linewidth}{>{\raggedright\arraybackslash}X*{6}{>{\centering\arraybackslash}X}}
\toprule
Method & VIF $\uparrow$ & SCD $\uparrow$ & $Q^{AB/F} \uparrow$  & MS-SSIM $\uparrow$ & FMI $\uparrow$ \\
\midrule
w/o \(\mathcal{L}_\text{ssim}\) & 0.7702 & 1.6773 & 0.5501 &  0.8260 & 0.8802 \\
w/o \(\mathcal{L}_\text{int}\) & 0.6673 & 1.8216 & 0.5448 &  0.8994 & 0.8804 \\
w/o \(\mathcal{L}_\text{tsxt}\) & 0.5235 & 1.8283 & 0.4604 &  0.9158 & 0.8862 \\
Ours & \textbf{0.7717} & \textbf{1.8314} & \textbf{0.5468}  & \textbf{0.9331} & \textbf{0.8858} \\
\bottomrule
\end{tabularx}
\vspace{-2.0em}
\end{table}

\section{Ablation Experiments}
\subsection{Structure Ablation}
To analyze the network architecture better, we conduct sets of ablation experiments, as shown in Table \ref{tab:ablation1}. To validate DVSS effectiveness, we perform two additional experiments: EVSS \cite{EVMamba} and replacing DVSS with Transformer (Case 1 and Case 2). EVSS produces satisfactory results, but it is less effective than DVSS in design due to DVSS's stronger feature extraction capability. Replacing DVSS with the Transformer network leads to a decline in metrics. The proposed DFFM significantly enhances fusion effect, as seen in Case 3 where its absence causes a decrease in \(Q^{AB/F}\). Removing the DFEM module (Case 4) shows similar results to not having DFFM. The Mamba module for cross-modal fusion is vital for effective information integration, as networks without Mamba module integration (Case 5) show degraded VIF task performance. Absence of CMFM leads to a decrease in MS-SSIM of fused images, indicating the importance of cross-domain integration for perceiving key goals and structures in fusion tasks.

\subsection{Loss Ablation}
We conduct ablation experiments on each loss function to evaluate its impact, as depicted in Table \ref{tab:ablation2}. Initially, we introduce SSIM loss (\(\mathcal{L}_\text{ssim}\)) to constrain the fusion network, preserving structural information in the source image. Furthermore, SSIM loss helps control the brightness of fusion results to some extent. Networks lacking structural constraints struggle to maintain optimal structure and strength information, leading to a decrease in MS-SSIM scores. Texture loss (\(\mathcal{L}_\text{text}\)) contributes significantly to retaining edge information in the fusion result, thereby enhancing clarity. Without texture loss, we observe significant decreases in \( Q^{AB/F} \). Excluding intensity loss (\(\mathcal{L}_\text{int}\)) from the joint loss function diminishes the visual impact of the fused image. Our model consistently outperforms other versions across all metrics, highlighting the optimal performance of our proposed loss function.
\subsection{Parameter Setting}
In the loss function $\mathcal{L} = \delta_1 \mathcal{L}_{\text{int}} + \delta_2 \mathcal{L}_{\text{text}} + \delta_3 \mathcal{L}_{\text{ssim}}$, the parameters $\delta_1$, $\delta_2$, and $\delta_3$ control the trade-off between intensity loss, texture loss, and structural similarity loss, respectively. To investigate their impacts, we conducted ablation experiments and adjusted the values based on the results. As shown in Table~\ref{tab:ablation_results}, different combinations of $\delta_1$, $\delta_2$, and $\delta_3$ affect performance across VIF, SCD, MS-SSIM, \(Q^{AB/F}\), and FMI metrics. Increasing $\delta_1$ improves VIF, indicating better detail preservation, but an excessively high $\delta_1$ may overemphasize fine details at the cost of overall structure. Increasing $\delta_2$ reduces SCD, mitigating texture distortion, but too large a $\delta_2$ can smooth out essential texture details. A larger $\delta_3$ enhances MS-SSIM, improving structural similarity, but an overly large $\delta_3$ can degrade the natural appearance of the image. We found that setting $\delta_1 = 20$, $\delta_2 = 10$, and $\delta_3 = 10$ achieves the best overall performance, with high scores in VIF, SCD, MS-SSIM, and FMI, indicating an optimal balance between preserving details, reducing texture distortion, and maintaining structural similarity. Due to space limitations, we present only one representative set of results. This combination of $\delta_1$, $\delta_2$, and $\delta_3$ provides the most balanced and effective fusion quality, as suggested by prior work in Refs. \cite{MACTFusion} and \cite{SwinFusion}.
\begin{figure*}[ht]
  \centering
  \vspace{-1em}
\includegraphics[width=\textwidth]{ 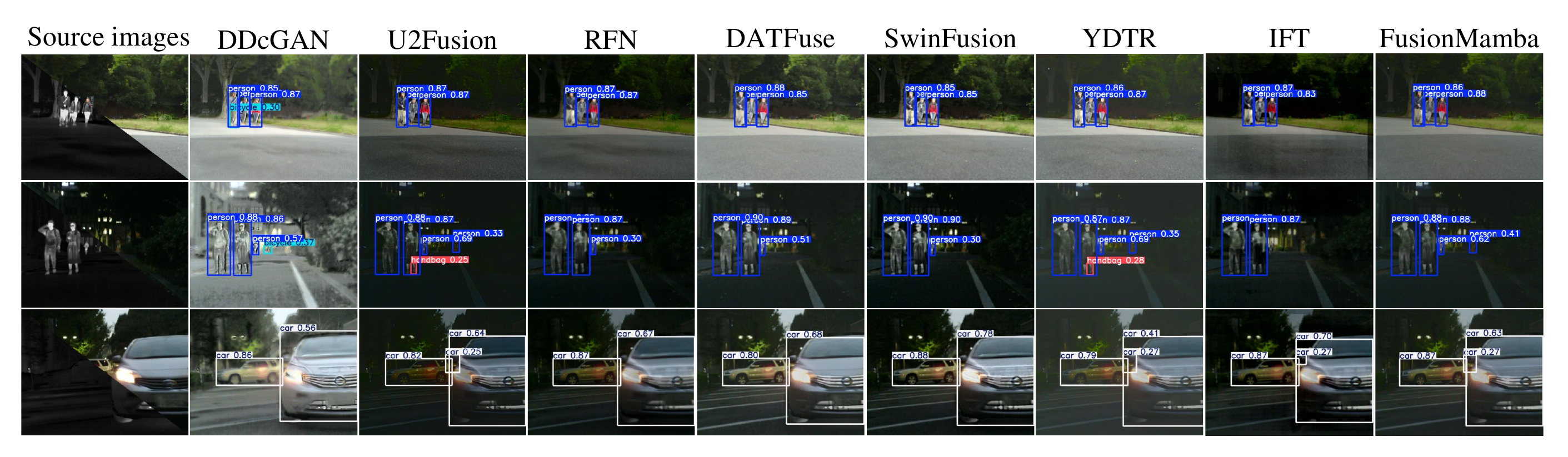}
\vspace{-1.5em}
  \caption{Visualization results of the object detection task using various methods on the MSRS \cite{MSRS} dataset.}
  \vspace{-0.3em}
  \label{fig:object-detection}
\end{figure*}

\begin{figure*}[ht]
  \centering
  \vspace{-2em}
  \includegraphics[width=\textwidth]{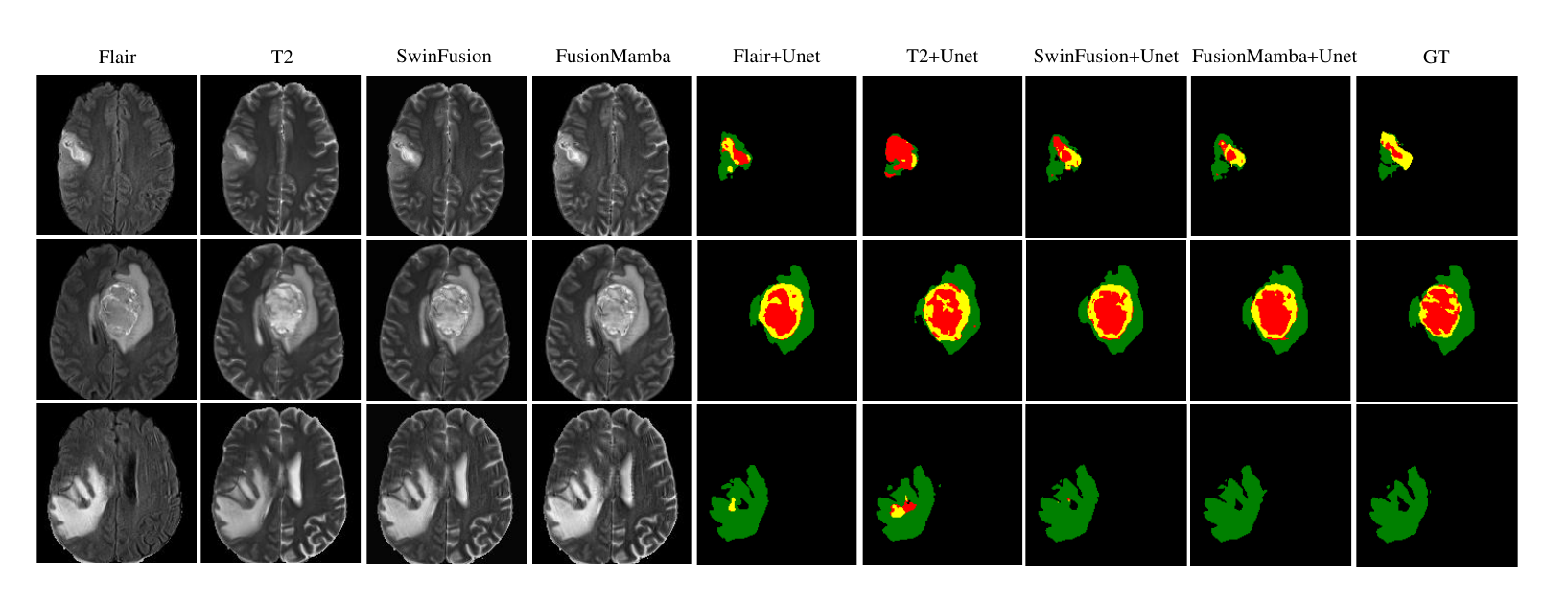}
  \vspace{-2.0em}
  \caption{From left to right: column 1 and column 2 are the MR-Flair and MR-T2 images, respectively;  column 3 and column 4 are the fusion results with SwinFusion and FusionMamba, respectively;  column 5 and column 6 are the single modality segmentation results of MR-Flair and MR-T2 with Unet, respectively;  column 7 and column 8 are the multimodal segmentation results of SwinFuion+Unet and FusionMamba+Unet, respectively. The MR-Flair and MR-T2 modality data are sourced from the BRATS19 dataset.}
  \label{fig:FusionMamba_seg}
  \vspace{-0.4em}
\end{figure*}
\section{Application of Downstream Tasks}
\subsection{Application to Object Detection}
To better evaluate the performance of FusionMamba and to explore its application in advanced vision tasks, we used IVF image fusion for target detection in our study. All fused images, as well as raw IR and visible images, are evaluated using pre-trained YOLOv5s \cite{YOLOv5} models. For a fair comparison, we kept the detection models on the fusion results of the seven state-of-the-art methods. The quantitative results in Table \ref{tab:detection} show that our FusionMamba outperforms the other methods in terms of precision (P), recall (R) and mean accuracy (mAP). Figure. \ref{fig:object-detection} illustrates the visualisation of the results of the various methods in the object detection task, showing that our FusionMamba outperforms in terms of overall recognition accuracy. This improvement is attributed to our specially designed dynamically enhanced Mamba fusion module and the Mamba framework, which effectively captures the semantic information and specific features of the source image.

\begin{table}[t]
\vspace{-0.5em}
\caption{Performance comparison of different models using various metrics on CT-MRI dataset.}
\resizebox{0.5\textwidth}{!}{%
\centering
\tiny
\label{tab:ablation_results}
\begin{tabular}{cccccccc}
\toprule
$\delta_{1}$ $\delta_{2}$ $\delta_{3}$ & VIF $\uparrow$ & SCD $\uparrow$ & $Q^{AB/F} \uparrow$  & MS-SSIM $\uparrow$& FMI $\uparrow$\\
\midrule
100\_20\_1 & 0.4707 & 1.5187 & 0.5075 & 0.9210 & 0.8546 \\
1\_10\_10 & 0.4267 & 1.3115 & 0.3733 & 0.8991 & 0.8470 \\
20\_1\_10 & 0.4218 & 1.2711 & 0.3593 & 0.8112 & 0.8493 \\
20\_20\_10 & 0.4791 & 1.4675 & 0.5859 & 0.9312 & 0.8507 \\
1\_10\_100 & 0.4217 & 1.2762 & 0.3593 & 0.8527 & 0.8311 \\
1\_1\_10 & 0.5567 & 1.2520 & 0.4530 & 0.8999 & 0.8611 \\
10\_1\_10 & 0.4218 & 1.2615 & 0.3589 & 0.8721 & 0.8447 \\
100\_1\_1 & 0.5743 & 1.5823 & 0.5910 & 0.9001 & 0.8600 \\
10\_1\_20 & 0.5293 & 1.3582 & 0.4390 & 0.8805 & 0.8675 \\
10\_1\_100 & 0.5270 & 1.2069 & 0.4309 & 0.9052 & 0.8587 \\
20\_10\_10 & \textbf{0.5750} & \textbf{1.5884} & \textbf{0.6429} & \textbf{0.9462} & \textbf{0.8624} \\
\bottomrule
\vspace{-2.5em}
\end{tabular}
}
\end{table}
\subsection{Application to brain tumor segmentation}
To illustrate the effectiveness of our FusionMamba approach in multimodal tumor fusion and subsequent segmentation tasks, we applied it to the MR-Flair and MR-T2 images from the Brats 2019 public dataset \cite{BRATS}. T2-weighted imaging (T2) in MRI is instrumental in highlighting fluid and soft tissues, with high signals observed in fluid and specific soft tissues, while bone and fat show low signals. This imaging technique is particularly beneficial for visualizing brain lesions, edema, cysts, and soft tissue anomalies. Fluid attenuated inversion recovery (Flair) imaging, on the other hand, is a specialized MRI technique that suppresses cerebrospinal fluid (CSF) signals, thereby enhancing the visibility of other tissues and abnormalities. It is especially valuable for identifying brain lesions, tumors, inflammation, and vascular irregularities. Given the complementary nature of these two modalities, we chose them for our experiments. The dataset includes 285 training sets and 49 test sets, each with a resolution of 240×240×155. In multi-label brain tumor segmentation, whole tumor (WT) represents the entire tumor region, tumor core (TC) denotes the tumor's core, and enhancing tumor (ET) refers to the enhancing tumor area, each used for more detailed analysis and differentiation of tumor components. The Dice coefficient measures the overlap between the model's segmentation and the ground truth, while sensitivity assesses the model's ability to correctly identify positive samples.

\begin{table}[t]
\centering
\vspace{-1.5em}
\caption{Object detection performance comparison across different methods. "IR" and "VIS" indicate the baseline result with IR and visible modality, respectively. P: precision, R: recall and mAP: mean accuracy.}
\tiny
\resizebox{0.5\textwidth}{!}{%
\begin{tabular}{lcccc}
\toprule
\multirow{2}{*}{Method} & \multirow{2}{*}{P} & \multirow{2}{*}{R} & \multicolumn{2}{c}{Object Detection (mAP)} \\ \cmidrule(lr){4-5}
                        &                    &                    & @.5           & @.5:.95           \\ \midrule
IR                      & 0.882              & 0.689              & 0.816         & 0.571             \\
VIS                     & 0.913              & 0.765              & 0.807         & 0.533             \\
DDcG \cite{DDCGAN}    & 0.898              & 0.834              & 0.904         & 0.637             \\
U2F \cite{U2Fusion}   & 0.908           & 0.801              & 0.921         & 0.611 \\
RFN \cite{RFN}             & 0.912           & 0.820              & 0.903         & 0.621    \\
DATFuse \cite{DATFuse}     & \textbf{0.924}   & 0.845             & 0.908         & 0.649             \\
SwinFusion \cite{SwinFusion}   & 0.910        & 0.749             & 0.890         & 0.577             \\
YDTR \cite{YDTR}        & 0.886              & 0.740              & 0.812         & 0.511             \\
IFT \cite{IFT}          & 0.892              & 0.815              & 0.882         & 0.597             \\
Ours  & 0.913   & \textbf{0.905}     & \textbf{0.945}&  \textbf{0.667}        \\ \bottomrule

\label{tab:detection}
\end{tabular}
 }
\vspace{-2.8em}
\end{table}

\begin{table*}[ht]
\centering
\tiny
\vspace{-0.5em}
\caption{Qualitative results of multimodal 2D segmentation are evaluated using the Unet, SwinFusion, and FusionMamba frameworks on the BRATS19 dataset. WT: whole tumor, TC: tumor core , ET: enhancing tumor. Dice coefficient measures segmentation overlap. Sensitivity evaluates the model's ability to identify positives.}
\label{tab:seg}
\begin{tabular}{ccccccc}
\toprule
\multirow{2}{*}{Method} & \multicolumn{3}{c}{Dice} & \multicolumn{3}{c}{Sensitivity} \\ \cmidrule(lr){2-4} \cmidrule(lr){5-7}
                        & WT     & TC     & ET     & WT      & TC      & ET      \\ \midrule
Unet \cite{Unet}+T2                 & 88.79  & 76.11  & 54.10  & 88.70   & 88.24   & 84.34   \\
Unet \cite{Unet}+Flair              & 83.34  & 78.63  & 62.19  & 84.77   & 85.79   & 80.64    \\
Unet \cite{Unet}+FlairT2            & 85.20  & 73.04  & 56.31  & 89.93   & 88.64   & 85.23   \\
SwinFusion \cite{SwinFusion}+Unet \cite{Unet}+FlairT2 & 85.64  & 78.21  &65.66   & 90.17   & 89.12  & 86.06   \\
FusionMamba+Unet \cite{Unet}+FlairT2   & \textbf{88.65}  & \textbf{80.30}  & \textbf{68.89}  & \textbf{90.80}   & \textbf{89.33}   & \textbf{86.77}   \\ \bottomrule
\end{tabular}
\vspace{-0.3em}
\end{table*}

For the segmentation task, we utilized the Unet \cite{Unet} model as the image segmentation network. Figure. \ref{fig:FusionMamba_seg} showcases three typical MR-Flair and MR-T2 image pairs, alongside their respective fusion results using SwinFusion and FusionMamba. The segmentation results are presented for both single-modal MR-Flair and MR-T2 images (Unet+MR-Flair and Unet+MR-T2, respectively), as well as for the fused images (SwinFusion+Unet and FusionMamba+Unet). As demonstrated in Table \ref{tab:seg}, we used the Dice coefficient and Sensitivity as metrics for quantitative analysis of the segmentation performance. The segmentation outcomes obtained using FusionMamba show a significant improvement in accuracy compared to both single-modal images and SwinFusion. This enhancement is primarily due to FusionMamba's capability to refine image details while preserving contrast, thereby bolstering the segmentation efficacy.

\section{Conclusion}
In conclusion, our study addresses the challenges of multimodal image fusion by proposing FusionMamba, a novel dynamic feature enhancement method integrated with the Mamba framework.  Our approach combines an improved efficient Mamba model with dynamic convolution and channel attention, enhancing both global modeling capabilities and local feature extraction.  We also introduce a dynamic feature fusion module (DFFM), which includes two dynamic feature enhancement modules (DFEM) and a cross-modality fusion Mamba module (CMFM), effectively enhancing texture, difference perception, and correlation between modalities.  Our FusionMamba method has demonstrated state-of-the-art performance in various multimodal image fusion tasks.  These results validate the generalization ability of our proposed method.  For future work, we aim to investigate the application of FusionMamba in real-time scenarios and deploy it on resource-constrained devices for practical implementation.  Furthermore, extending our evaluation to more diverse datasets and benchmarking against emerging fusion methods would provide a comprehensive understanding of FusionMamba's capabilities.

\begin{small}

\vspace{.3in} \noindent \textbf{Data Availability Statement:}
The datasets supporting this study’s findings are publicly accessible and include three infrared and visible image datasets: KAIST \cite{KAIST} \footnote{\url{https://sites.google.com/site/pedestrianbenchmark/home}}, MSRS \cite{MSRS} \footnote{\url{https://github.com/Linfeng-Tang/MSRS}}, TNO \cite{TNO} \footnote{\url{https://figshare.com/articles/dataset/TNO_Image_Fusion_Dataset/1008029}}, and RoadScene \cite{RoadScene} \footnote{\url{https://github.com/hanna-xu/RoadScene}}. Additionally, three medical image datasets (covering CT-MRI, PET-MRI, and SPECT-MRI tasks) were sourced from the Harvard Public Medical Dataset\footnote{\url{http://www.med.harvard.edu/AANLIB/home.html}}, alongside a biomedical image dataset of \textit{Arabidopsis thaliana} cells from the John Innes Centre \cite{GFPPC} \footnote{\url{http://data.jic.ac.uk/Gfp/}}. The MSRS dataset \cite{MSRS} was used for target detection experiments, and the BraTS2019 dataset \cite{BRATS} \footnote{\url{https://www.med.upenn.edu/cbica/brats-2019/}} provided multimodal medical image segmentation data. While data from institutions such as Wuhan University, the Netherlands Organisation for Applied Scientific Research, Harvard Medical School, the John Innes Centre, and the Center for Biomedical Image Computing and Analytics are available, their use is restricted due to licensing agreements and is not publicly accessible. However, the data can be made available upon reasonable request and with permission from the relevant third-party institutions. The code for FusionMamba is
available at https://github.com/millieXie/FusionMamba.

\vspace{.3in} \noindent \textbf{Abbreviations:}
CMFM: cross modality fusion Mamba module; DFFM: dynamic
feature fusion Module; DFEM: dynamic feature enhancement module; MIF: medical image fusion.

\vspace{.3in} \noindent \textbf{Competing Interests:}
The authors declare no competing interests.

\vspace{.3in} \noindent \textbf{Authors' Contributions:}
Xinyu Xie is the first author for main idea generation and implementation. Yawen Cui is the second author for parital experiments and draft writing. Tao Tan is the co-supervisor of this work, and provides comprehensive guidance and draft writing. Xubin Zheng is in charge of ablation experiments and corresponding tables and figures. Zitong Yu is the project leader who organizes the process of discussing ideas, experimenting and writing.

\end{small}

\begin{acknowledgements}
Not applicable.

\vspace{.3in} \noindent \textbf{Fundings:}
This work was supported by National Natural Science Foundation of China (No. 62306061), Guangdong Basic and Applied Basic Research Foundation (No. 2023A1515140037), Macao Polytechnic University Grant (RP/FCA-05/2022) and Macao Polytechnic University Grant (RP/FCA-15/2022).
\end{acknowledgements}

\bibliographystyle{unsrt}
\bibliography{reference}

\end{document}